\documentclass[11pt]{article}

\usepackage[preprint]{acl}

\usepackage{amsmath}
\usepackage{amssymb}
\usepackage{amsthm}         

\usepackage{times}
\usepackage{latexsym}

\usepackage{booktabs}
\usepackage[table]{xcolor}
\definecolor{light-gray}{gray}{0.90}
\usepackage{multirow}
\usepackage{enumitem}
\usepackage{bbm}

\usepackage[T1]{fontenc}

\usepackage[utf8]{inputenc}

\usepackage{microtype}

\usepackage{inconsolata}

\usepackage{graphicx}
\usepackage{subcaption}

\usepackage{xspace}
\newcommand{\model}{\textsc{TAR}\xspace}

\title{Preserving Reasoning Diversity in Reinforcement Learning with Verifiable Rewards via Adaptive Replay}

\author{Wen-Zhi Li \quad Sainyam Galhotra \quad Claire Cardie \\
  Cornell University \\
  \texttt{\{wenzhi, sg, cardie\}@cs.cornell.edu} \\}

\begin{document}
\maketitle
\begin{abstract} 
Reinforcement Learning with Verifiable Rewards (RLVR) improves the reasoning capabilities of large language models, but often reduces the diversity of their reasoning behaviors. We study this phenomenon from the perspective of finite-sample learning dynamics and show that limited on-policy rollouts can progressively suppress semantically distinct but equally rewarded reasoning modes. Once a mode becomes sufficiently under-sampled, standard on-policy RLVR rarely recovers it. Motivated by this observation, we propose \model, a Targeted Adaptive Replay framework that preserves diverse correct reasoning behaviors already present in the base policy. \model constructs a compact memory bank of diverse base-model solutions and selectively replays them through an adaptive mechanism that activates only when stored behaviors are being suppressed during RLVR. Across six mathematical reasoning benchmarks and two Qwen2.5 model scales, \model consistently achieves a better balance between reasoning performance and diversity preservation than existing diversity-preserving post-training methods, while substantially reducing the forgetting of correct base-model solutions. Code is available at \url{https://github.com/wenzhilics/tar}.
\end{abstract}

\section{Introduction}

Reinforcement Learning with Verifiable Rewards (RLVR) has emerged as the de facto paradigm for equipping Large Language Models (LLMs) with advanced reasoning capabilities, particularly in mathematical and coding domains~\cite{jaech2024openai, guo2025deepseek, yang2025qwen3}. The standard formulation optimizes a KL-regularized objective over model-generated responses with respect to a verifiable reward function, with Group Relative Policy Optimization (GRPO)~\cite{shao2024deepseekmath} becoming the dominant training recipe.
Despite these advances, recent studies suggest that RLVR post-training does not necessarily induce genuinely new capabilities; rather, it tends to amplify high-reward regions already present in the base model's prior distribution~\cite{yue2025does, wu2025invisible}.
This behavior can lead to reduced output diversity and insufficient exploration, ultimately resulting in mode collapse.

From a theoretical perspective, recent studies show that for any two responses with equal reward, their probability ratio in the optimum is identical to that in the reference policy~\cite{korbak2022reinforcement, gx2025kl, chen2025retaining}. Therefore, the objective itself does not introduce preference against any mode. This raises a natural question: \textit{if equally rewarded modes are not explicitly distorted by the objective, why do some semantically valid reasoning modes nevertheless disappear during RLVR in practice?}

We analyze RLVR from the perspective of finite-sample learning dynamics and show that stochastic fluctuations induced by limited on-policy rollouts can accumulate across updates, progressively shifting probability mass toward a subset of reasoning modes while others become increasingly underrepresented. This analysis identifies the coverage of distinct correct reasoning modes as a key factor behind this instability, suggesting a direct design principle: mitigating mode collapse requires increasing such coverage during training.

Existing approaches encourage exploration in RLVR mainly by modifying the learning signal over on-policy rollouts to introduce explicit diversity rewards~\cite{he2025rewarding, cheng2026reasoning, yao2025diversity, chen2026posttraining}. However, such methods can only act on behaviors already represented in the sampled rollouts, providing no direct mechanism to recover a sufficiently under-sampled mode.

\begin{figure}[!t]
    \centering
    \includegraphics[width=\linewidth]{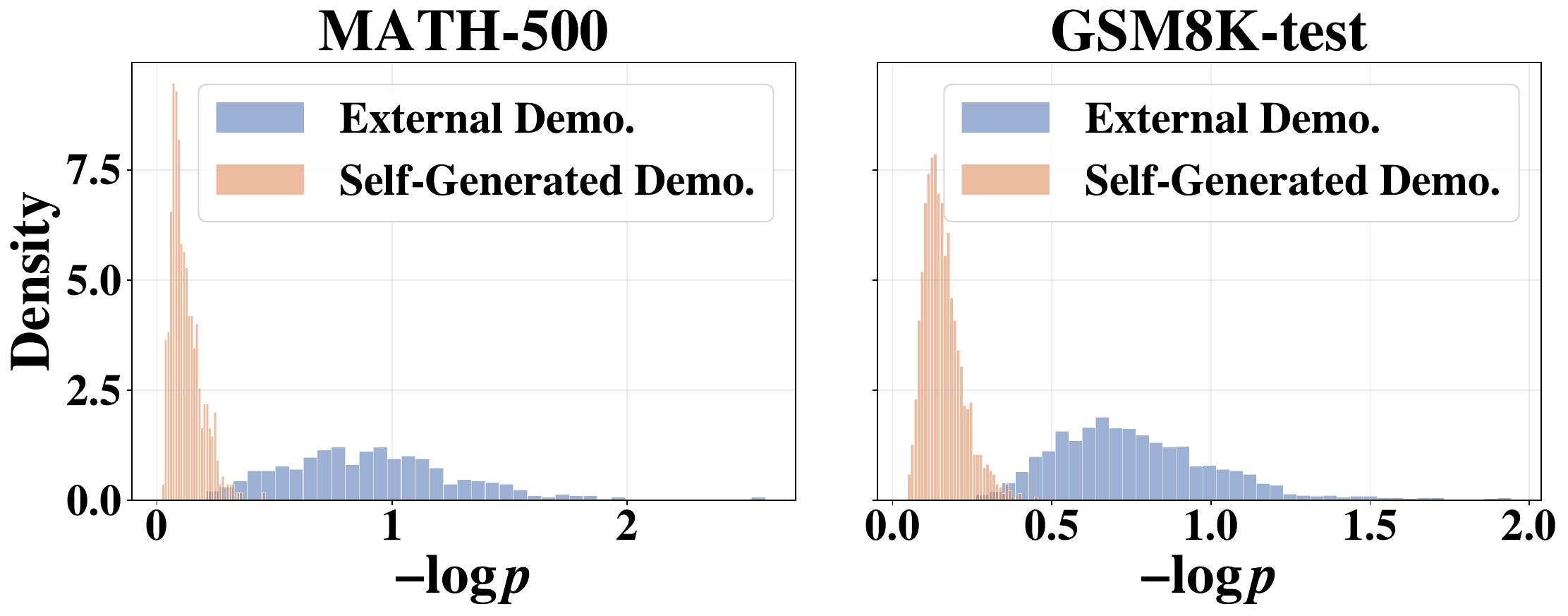}
    \caption{Negative log-likelihood distribution of ground truth external expert demonstrations and internal LLM self-generated demonstrations with Qwen2.5-0.5B-Instruct.
    }\label{fig:discrenpency}
\end{figure}

A straightforward way to improve such coverage is to increase the number of on-policy rollouts. However, this is often computationally prohibitive, as rollout generation constitutes a major bottleneck in GRPO training~\cite{vllm}. Another possibility is to introduce diverse demonstrations from human experts or stronger LLMs~\cite{luffy, yuan2025more}. Yet such external trajectories can be substantially out-of-distribution under the current policy, which can weaken their effectiveness as positive training signals~\cite{zhou2025expo}, as illustrated in Figure~\ref{fig:discrenpency}.
These observations motivate a different strategy: rather than repeatedly generating more on-policy rollouts or relying on external demonstrations, we can reuse diverse correct reasoning trajectories already intrinsically in the base model itself.

To this end, we propose Targeted Adaptive Replay (\model), an RLVR framework designed to preserve diverse correct reasoning behaviors already present in the base policy and selectively restore them during training.
Specifically, we first construct a compact memory bank by oversampling base-model trajectories, filtering for correctness, and selecting semantically diverse prototypes.
We then introduce a gated adaptive replay mechanism that selectively activates supervision when a stored trajectory has become less likely under the current policy than under the base policy, while concentrating the corrective update on its most suppressed tokens.
This allows the memory bank to act as a targeted off-policy corrective signal for disappearing reasoning modes, rather than a persistent imitation objective.

Empirically, across six mathematical reasoning benchmarks and two Qwen2.5 model scales, \model consistently achieves a favorable balance between reasoning performance and diversity preservation compared with existing diversity-preserving methods. Further analysis shows that \model substantially reduces the forgetting of correct reasoning behaviors inherited from the base model. Together, these results demonstrate that targeted adaptive replay can preserve a broader range of reasoning behaviors without preventing effective RL adaptation.

\section{Preliminary}

An LLM can be modeled as a policy $y\sim\pi_{\theta}(\cdot|x)$, where the output response $y$ is generated given the input prompt $x$. In RLVR for LM post-training, the objective is to maximize the KL-regularized reward with respect to responses generated by the LM as 
\begin{equation}\label{eq:rlobj}\small
    J_{\text{RL}}(\theta) := \mathbb{E}_{y \sim \pi_\theta(\cdot \mid x)}\left[R(x, y)\right] - \beta \mathrm{KL}\left[\pi_\theta \| \pi_{\mathrm{ref}}\right],
\end{equation}
where $R(x, y) \in \{0, 1\}$ is a binary reward the response $y$ gets given prompt $x$, and $\pi_{\mathrm{ref}}$ is the initial policy.
In practice, recent RLVR methods commonly adopt Group Relative Policy Optimization (GRPO) \cite{shao2024deepseekmath}, which estimates the relative advantage of sampled responses within a group. Specifically, given a prompt $x$, GRPO samples a group of responses $\{y_i\}_{i=1}^G \sim \pi_\theta(\cdot|x)$ and computes the normalized advantage as
\begin{equation}\small
    \hat{A}_i = \frac{R(x,y_i)-\mathrm{mean}(\{R(x,y_j)\}_{j=1}^G)}
    {\mathrm{std}(\{R(x,y_j)\}_{j=1}^G)}.
\end{equation}
The policy is then optimized using PPO~\cite{ppo}-style clipped policy gradients with KL regularization toward the reference policy. 

Theoretically, the optimal policy for Eq.~\eqref{eq:rlobj} is $\pi^{*}\propto \pi_{\mathrm{ref}}\exp\left( R(x, y) / \beta \right)$~\cite{korbak2022reinforcement}, which implies there is no inherent incentive to distort the correct responses' probability ratio after RLVR. 
However, recent studies have observed a ``mode collapse'' effect in RLVR training empirically: LLMs post-trained with RLVR generate less diverse outputs~\cite{zhang2025noveltybench, yue2025does}. 
To study how such imbalance can arise during training, we analyze the evolution of probability mass across semantically distinct correct reasoning modes.
For clarity, we focus on the two-mode case.
Let the correct responses for a prompt $x$ be partitioned into disjoint semantic modes
$\{\mathcal{M}_1,\mathcal{M}_2\}$.
For a policy $\pi_t$ at training step $t$, we define the probability mass of mode $\mathcal{M}_k$ as $P_{k,t}:=\pi_t(\mathcal{M}_k)=\sum_{y\in\mathcal{M}_k}\pi_t(y)$.
\begin{figure*}[!t]
    \centering

    \begin{subfigure}[t]{0.42\linewidth}
        \centering
        \includegraphics[width=\linewidth]{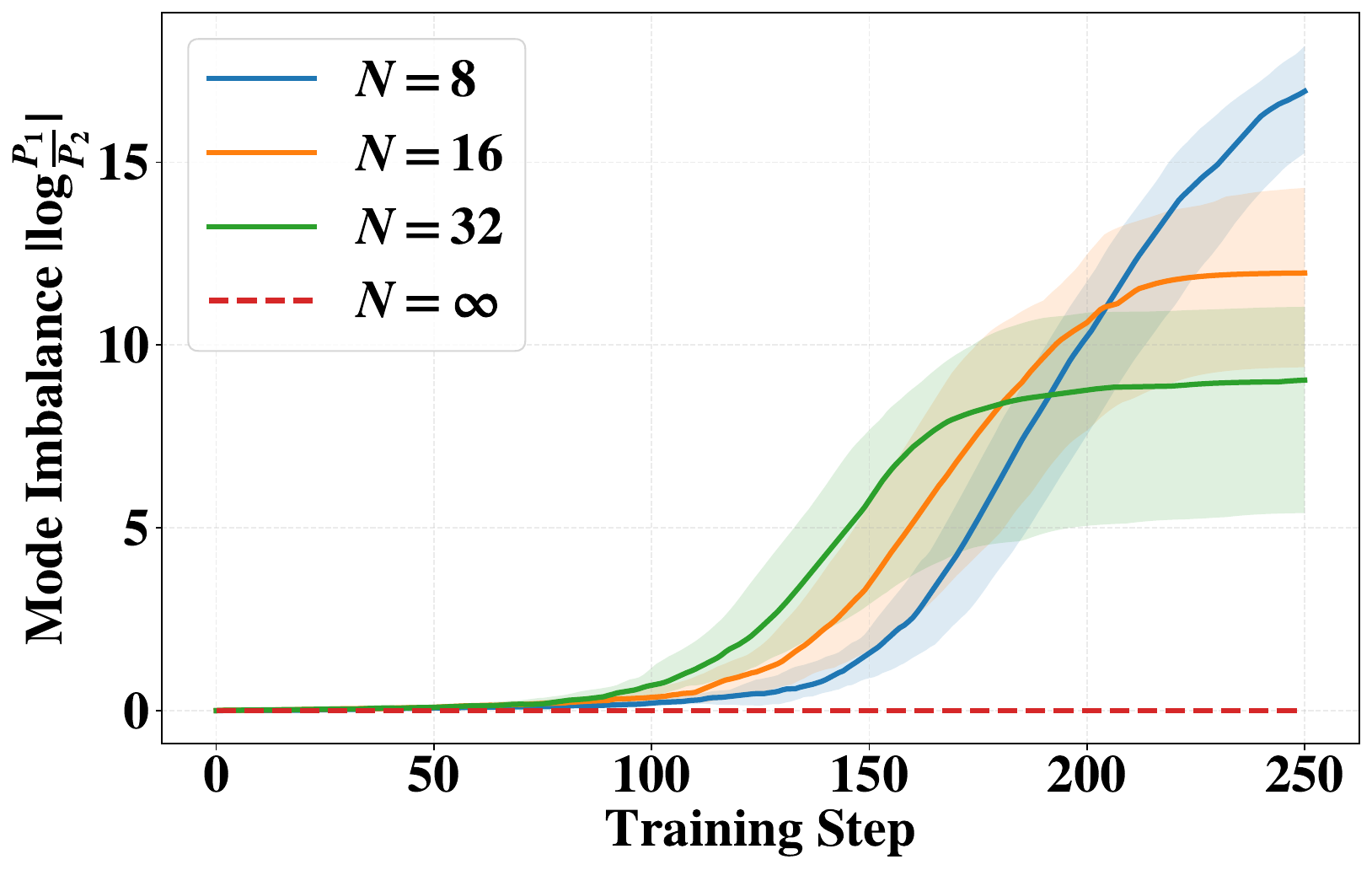}
        \caption{}\label{fig:symmetry_breaking}
    \end{subfigure}
    \qquad
    \begin{subfigure}[t]{0.42\linewidth}
        \centering
        \includegraphics[width=\linewidth]{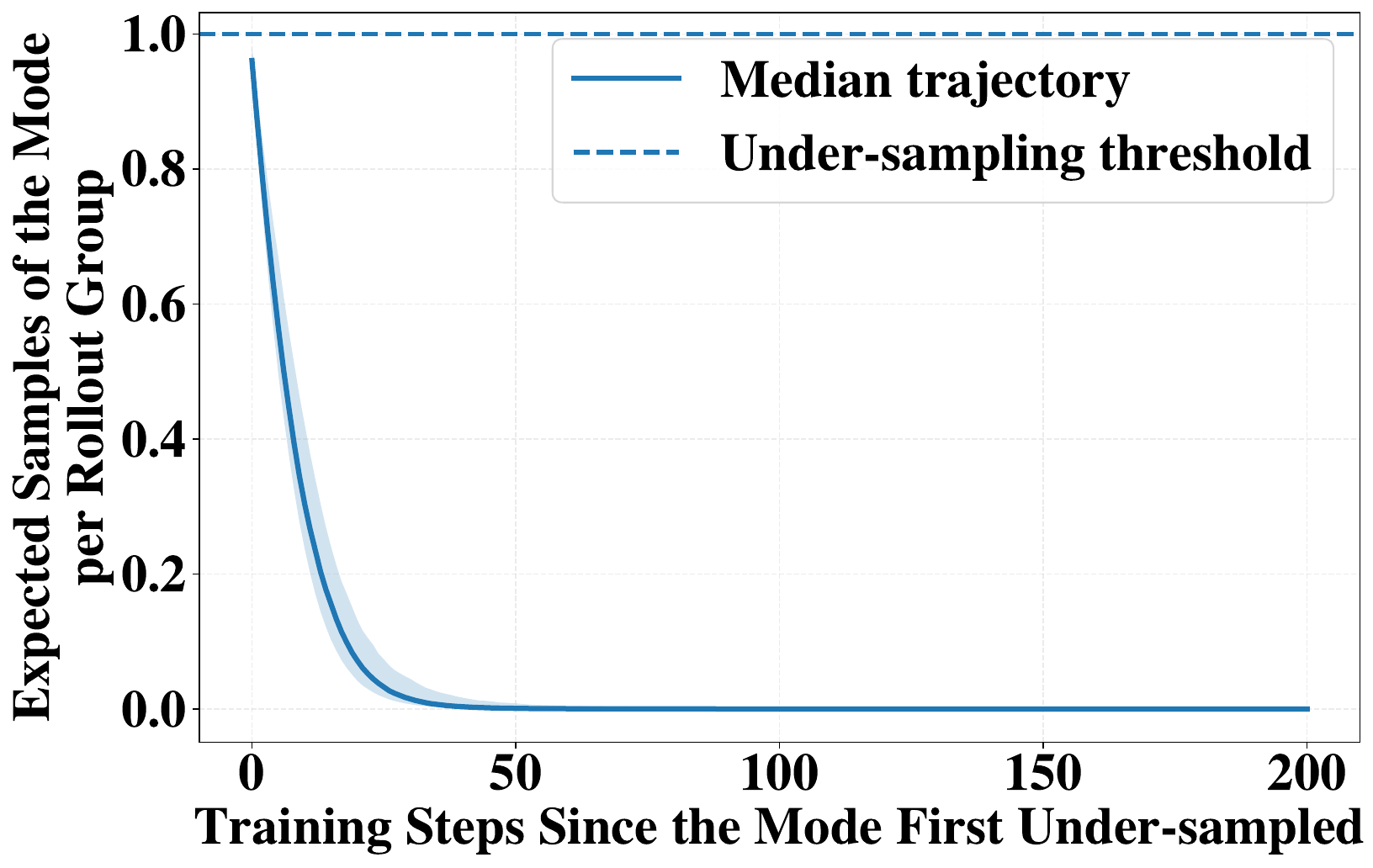}
        \caption{}\label{fig:mode_recovery}
    \end{subfigure}

\caption{
Toy analysis of mode imbalance and recovery under finite-sample RLVR.
(a) Mode imbalance between two equally rewarded modes for different rollout-group sizes.
(b) Mode coverage over the 200 updates after the first crossing of $C_{k,t}<1$, aligned at the crossing step. Shaded regions denote interquartile ranges.
}
\end{figure*}

\section{Why Do Reasoning Modes Disappear?}
\label{sec:analysis}

We study the dynamics in a controlled discrete analogue of RLVR.
The policy is a categorical distribution over 300 response states, where each state represents a complete response.
Two disjoint subsets of responses represent correct semantic modes, $\mathcal{M}_1$ and $\mathcal{M}_2$.
Responses in these two modes are assigned a higher initial logit of 7, while all remaining responses are assigned logit 2, producing high-probability correct modes over a low-probability background.
All responses in the two modes receive reward 1, and all remaining responses receive reward 0.
At each training step, we sample $N$ responses from the current categorical policy and update its logits using the GRPO-style objective.

Our analysis asks two questions:
(1) can finite on-policy sampling create imbalance between otherwise equivalent modes, and
(2) once a mode becomes sufficiently under-sampled, can ordinary on-policy RLVR recover it?

\subsection{Finite Sampling Breaks Mode Symmetry}
\label{sec:symmetry_breaking}

We first ask whether finite sampling alone can destabilize two equally valid modes.
We set $|\mathcal{M}_1|=|\mathcal{M}_2|=20$, so that $P_{1,0}=P_{2,0}$ exactly. We vary the rollout-group size $N\in\{8,16,32\}$ and run 50 independent trials for each setting. 
We additionally show the noise-free $N\to\infty$ limit as a reference, where the two symmetric modes remain equally weighted.

Figure~\ref{fig:symmetry_breaking} shows that finite-sample training progressively separates the two mode masses as $|\log\frac{P_{1,t}}{P_{2,t}}|$ increases, despite the two modes having identical reward and identical initial probability.
The imbalance is also generally more severe for smaller rollout groups.
This experiment isolates a simple source of mode imbalance: finite on-policy sampling can spontaneously break the symmetry between equally valid reasoning modes.

\subsection{Under-Sampled Modes Rarely Recover}
\label{sec:mode_recovery}

We next ask whether a mode can recover once it becomes sufficiently under-sampled.
For a mode $\mathcal{M}_k$ with probability mass $P_{k,t}$, define $C_{k,t}:=NP_{k,t}$ as the expected number of samples from that mode in one rollout group.
We call a mode \emph{under-sampled} when $C_{k,t}<1$, i.e., when it contributes fewer than one sample per rollout group in expectation.

To cover different initial mode asymmetries, we fix $|\mathcal{M}_1|=20$ and vary $|\mathcal{M}_2|\in\{20,10,5\}$. We evaluate two KL coefficients, $\beta\in\{0.01,0.05\}$, and set group size to $N\in\{8,16,32\}$.
For each run, we identify the first mode whose expected coverage falls below one within the first 1,000 training steps and track its subsequent evolution for 200 updates.
Across all configurations, we run 900 trials in total.
We consider the mode recovered if its expected coverage returns to $C_{k,t}\geq 1$ within this fixed observation window.

Overall, 832 out of 900 runs ($92.4\%$) enter the under-sampled regime within the first 1,000 training steps.
Among these runs, only $2.6\%$ recover to $C_{k,t}\geq 1$ within the subsequent 200 updates.
Figure~\ref{fig:mode_recovery} aligns trajectories at their first threshold crossing and shows this fixed 200-step observation window.
Rather than rebounding toward $C_{k,t}=1$, the median trajectory rapidly decays toward zero and remains near zero throughout the window.

Together, these experiments reveal a simple failure mode of finite-sample RLVR:
finite rollout sampling creates imbalance between equally rewarded modes. Once a mode becomes sufficiently under-sampled, subsequent on-policy training rarely restores it.

\section{\model}

As shown in Section~\ref{sec:analysis}, once a reasoning behavior becomes sufficiently under-sampled, it may no longer appear often enough in on-policy rollouts to receive an effective corrective signal. We therefore introduce an off-policy replay mechanism that can reinforce a stored trajectory independently of whether it is sampled by the current policy and propose \model.
\model constructs a memory bank that stores diverse rollout prototypes from the base policy, and then incorporates the memory into RLVR training. The method is also consistent with the findings of~\cite{wu2025invisible}, which suggests that RLVR should be coupled with off-policy mechanisms that inject probability mass into underrepresented regions of the reasoning space to improve exploration and reasoning diversity.

\subsection{Memory Bank Construction}

The goal of the memory bank is to collect correct rollouts with high diversity from the base policy in an off-policy manner. Compared with on-policy oversampling during RL training, rollouts generated by the base policy better align with the original model distribution while also avoiding the substantial computational cost of repeated on-policy exploration. To achieve this, we adopt a RAFT~\cite{dong2023raft}-style rejection sampling procedure combined with furthest point search (FPS)~\cite{fps}, due to its simplicity to select well-separated trajectories.

Specifically, for each training question $x_i$, we sample $N$ rollouts
$\{y_i^{(1)}, y_i^{(2)}, \cdots, y_i^{(N)}\}$
from the base policy. We retain only the $M$ rollouts that receive a positive reward,
$\{y_i^{(1)}, y_i^{(2)}, \cdots, y_i^{(M)}\}$,
and encode them using an embedding model $E(\cdot)$ to obtain
$\{E(y_i^{(1)}), E(y_i^{(2)}), \cdots, E(y_i^{(M)})\}$.
We then perform FPS in the embedding space and retain at most $k$ diverse rollouts for each question.

It is worth noting that many alternative approaches can also be used to construct the memory bank, as long as the collected rollouts are both correct and diverse. Examples include in-context regeneration~\cite{zhang2025noveltybench} and regeneration from fork tokens~\cite{wang2025beyond}. We leave the exploration of these alternatives to future work.

\subsection{Training Recipe}

After constructing the memory bank, we next consider how to incorporate its stored trajectories into RLVR training. Since each rollout in the memory bank is already a verified correct trajectory, we use supervised fine-tuning (SFT) as a direct way to increase its likelihood.

However, directly applying the SFT loss on all stored correct rollouts introduces a different mismatch with the objective. Our goal is not to continuously imitate the finite memory bank, but to preserve diverse behaviors from the base policy that would otherwise be suppressed during RLVR. A stored trajectory does not require corrective replay if RLVR has not reduced its likelihood relative to the base policy. Continuing to apply SFT in this case introduces unnecessary imitation pressure and may over-concentrate the policy on the finite replay bank. Therefore, replay should be activated only when a stored behavior has become less likely than it was under the base policy.

To this end, we introduce a base-policy-aware adaptive replay. Let $\tau=(x,y_{1:T})$ denote a stored rollout from the memory bank, where $y_t$ is the $t$-th token. We compare the token-level log-probability under the base policy $\pi_{\mathrm{base}}$ and the current policy $\pi_\theta$:
\begin{equation}\small\label{eq:delta}
    \delta_t(\tau;\theta)
    =
    \log \pi_{\mathrm{base}}(y_t \mid x,y_{<t})
    -
    \log \pi_\theta(y_t \mid x,y_{<t}).
\end{equation}
A positive $\delta_t$ indicates that the token is preferred more by the base policy than by the current policy, suggesting that the corresponding reasoning has been suppressed during RL training.

We then define the average trajectory-level discrepancy as
\begin{equation}\small
    b(\tau;\theta)
    =
    \frac{1}{T}
    \sum_{t=1}^{T}
    \delta_t(\tau;\theta),
\end{equation}
which is a trajectory-level measure of suppression. We convert the log-likelihood discrepancy into a one-sided relative likelihood deficit using
\begin{equation}\small
\phi(x)=\max\left(1-\exp(-x),0\right),
\end{equation}
which assigns zero weight when no suppression is detected, increases smoothly with the magnitude of suppression, and remains bounded for large discrepancies.

Based on this discrepancy, we define a rollout-level gate
\begin{equation}\small
    g(\tau;\theta)=\phi(b(\tau;\theta)),
\end{equation}
which determines whether a stored trajectory should receive replay supervision. If $b(\tau;\theta)\leq 0$, the trajectory has not become less likely on average than under the base policy, so no corrective replay is applied. If $b(\tau;\theta)>0$, its average log-likelihood has decreased relative to the base policy, and the replay loss is activated to restore this under-preserved behavior.

The rollout-level gate determines whether replay is needed, but a suppressed trajectory need not be uniformly suppressed at every token. We therefore further localize the corrective signal using token-level weighting:
\begin{equation}\small
w_t(\tau;\theta)=\phi(\delta_t(\tau;\theta)).
\end{equation}
Thus, only token transitions suppressed relative to the base policy contribute to the replay loss.

Given a replay mini-batch $\mathcal{B}_{\mathrm{rep}}$ sampled from the memory bank, we define the adaptive replay loss as
\begin{equation}\small
\begin{split}
\mathcal{L}_{\mathrm{rep}}(\theta) = {} & \frac{1}{|\mathcal{B}_{\mathrm{rep}}|}
\sum_{\tau\in\mathcal{B}_{\mathrm{rep}}} \mathrm{sg}[g(\tau;\theta)] \\
& \cdot\frac{\sum_{t=1}^{T_\tau}\mathrm{sg}[w_t(\tau;\theta)]\left[-\log \pi_\theta(y_t \mid x,y_{<t})\right]}
{\epsilon+\sum_{t=1}^{T_\tau}\mathrm{sg}[w_t(\tau;\theta)]},
\end{split}
\end{equation}
where $\mathrm{sg}[\cdot]$ denotes stop-gradient and $\epsilon=10^{-8}$.
We average the replay objective over stored trajectories, treating each retained rollout as an independent training instance. Thus, when the memory bank contains multiple diverse trajectories for the same prompt, each trajectory contributes its own corrective signal.

The final training objective combines the on-policy GRPO objective with the adaptive replay regularizer:
\begin{equation}\small
    \max_\theta
    \quad
    J_{\mathrm{GRPO}}(\theta)
    -
    \lambda_{\mathrm{rep}}
    \mathcal{L}_{\mathrm{rep}}(\theta),
\end{equation}
where $\lambda_{\mathrm{rep}}$ controls the strength of replay regularization. 
Overall, the memory bank determines which correct and diverse behaviors are worth preserving, while the adaptive replay mechanism determines when correction is needed and where it should be applied. The resulting replay objective therefore acts as a selective off-policy corrective signal rather than a persistent imitation objective.

\section{Experiments}

\subsection{Experimental Setup}

\begin{table*}[t]
\centering

\resizebox{0.9\textwidth}{!}{%
\begin{tabular}{cl|cccccc|c}
\toprule
&
\textbf{Model}
&
\textbf{GSM8K}
&
\textbf{MATH-500}
&
\textbf{AMC}
&
\textbf{College Math}
&
\textbf{Minerva}
&
\textbf{OlympiadBench}
&
\textbf{Avg.}
\\
\midrule
\midrule

\multirow{16}{*}{\rotatebox{90}{\textbf{Qwen-2.5-0.5B}}}
& Base
& 46.9 / 77.6
& 30.8 / 62.0
& 7.5 / 35.0
& 27.4 / \textbf{45.0}
& 5.5 / 18.4
& 7.1 / 27.4
& 20.9 / 44.2
\\

& GRPO
& \textbf{52.7} / 75.5
& 36.6 / 62.0
& 10.0 / 40.0
& 30.6 / 43.3
& \textbf{8.8} / 22.8
& 9.8 / \textbf{28.9}
& 24.8 / 45.4
\\

\cmidrule(lr){2-9}
& \multicolumn{8}{c}{\cellcolor{light-gray}\textbf{On-Policy Diversity}}
\\
\cmidrule(lr){2-9}

& Unlikely
& 35.1 / 73.0
& 24.6 / 56.8
& 5.0 / 27.5
& 23.3 / 42.5
& 4.0 / 21.0
& 6.5 / 24.7
& 16.4 / 40.9
\\

& R1-Zero-Div
& 5.4 / 22.1
& 7.6 / 25.8
& 5.0 / 25.0
& 7.2 / 17.1
& 3.7 / 10.3
& 3.1 / 13.0
& 5.3 / 18.9
\\

& DQO
& 30.8 / 60.1
& 22.6 / 41.2
& 2.5 / 17.5
& 27.0 / 42.3
& 2.6 / 7.7
& 2.4 / 10.2
& 14.7 / 29.8
\\

& DPH-RL
& 50.7 / 77.3
& 34.8 / 60.4
& 12.5 / 40.0
& 30.4 / 44.8
& 7.4 / 21.7
& 8.4 / 26.2
& 24.0 / 45.1
\\

\cmidrule(lr){2-9}
& \multicolumn{8}{c}{\cellcolor{light-gray}\textbf{Offline Replay}}
\\
\cmidrule(lr){2-9}

& \multicolumn{1}{>{\columncolor{light-gray}}l}{\textbf{Mixed-Policy}}
& \multicolumn{7}{l}{}
\\

& MixPolicy
& 49.1 / 74.5
& 34.8 / 60.8
& 2.5 / 32.5
& 29.6 / 43.8
& 6.6 / 18.8
& 8.7 / 27.3
& 21.9 / 43.0
\\

& LUFFY
& 50.1 / \textbf{78.0}
& 35.2 / 61.2
& 10.0 / 37.5
& 29.7 / 44.3
& 7.0 / 18.8
& 7.4 / 26.7
& 23.2 / 44.4
\\

& CalibRL
& 50.9 / 76.6
& \textbf{37.6} / 61.8
& 10.0 / 37.5
& \textbf{31.2} / 44.4
& 8.5 / \textbf{23.2}
& 10.5 / 27.4
& 24.8 / 45.2
\\

& \multicolumn{1}{>{\columncolor{light-gray}}l}{\textbf{Supervised}}
& \multicolumn{7}{l}{}
\\

& SFT
& 48.2 / 75.8
& 32.8 / 61.2
& 10.0 / 37.5
& 28.0 / 43.7
& 6.3 / 17.3
& 7.9 / 25.6
& 22.2 / 43.5
\\

& RL+SFT
& 46.9 / 76.0
& 34.4 / 60.4
& 5.0 / \textbf{47.5}
& 28.6 / 43.8
& 7.0 / 17.6
& 8.0 / 26.7
& 21.7 / 45.3
\\

& \model
& 50.7 / 77.7
& 37.2 / \textbf{62.4}
& 12.5 / \textbf{47.5}
& 30.4 / 44.5
& 7.7 / 20.6
& \textbf{10.7} / 28.7
& \textbf{24.9} / \textbf{46.9}
\\

\midrule
\midrule

\multirow{16}{*}{\rotatebox{90}{\textbf{Qwen-2.5-1.5B}}}
& Base
& 72.6 / \textbf{92.3}
& 50.6 / 76.0
& 25.0 / 62.5
& 37.5 / 51.4
& 15.4 / 36.8
& 19.1 / 40.3
& 36.7 / 59.9
\\

& GRPO
& 74.4 / 91.1
& 53.8 / 78.8
& 20.0 / 62.5
& 40.0 / 50.3
& 18.4 / 37.1
& 20.7 / 41.2
& 37.9 / 60.2
\\

\cmidrule(lr){2-9}
& \multicolumn{8}{c}{\cellcolor{light-gray}\textbf{On-Policy Diversity}}
\\
\cmidrule(lr){2-9}

& Unlikely
& 51.1 / 92.0
& 34.2 / 75.2
& 17.5 / 57.5
& 17.9 / 45.0
& 9.6 / 31.3
& 12.7 / 37.8
& 23.8 / 56.5
\\

& R1-Zero-Div
& 73.9 / 92.1
& 54.4 / 76.2
& 27.5 / 60.0
& 39.6 / \textbf{52.1}
& 16.2 / 33.8
& 19.1 / 40.3
& 38.5 / 59.1
\\

& DQO
& 72.9 / \textbf{92.3}
& 44.4 / 73.8
& 15.0 / 50.0
& 35.6 / 50.7
& 14.0 / 34.6
& 16.1 / 37.2
& 33.0 / 56.4
\\

& DPH-RL
& 73.8 / 92.0
& 55.2 / 77.8
& 25.0 / 52.5
& 39.6 / 50.5
& 18.4 / 36.4
& 19.1 / 39.4
& 38.5 / 58.1
\\

\cmidrule(lr){2-9}
& \multicolumn{8}{c}{\cellcolor{light-gray}\textbf{Offline Replay}}
\\
\cmidrule(lr){2-9}

& \multicolumn{1}{>{\columncolor{light-gray}}l}{\textbf{Mixed-Policy}}
& \multicolumn{7}{l}{}
\\

& MixPolicy
& 74.6 / 90.2
& 55.6 / 76.2
& \textbf{30.0} / \textbf{67.5}
& 40.0 / 49.6
& 18.0 / 34.2
& 19.7 / 39.6
& 39.7 / 59.6
\\

& LUFFY
& 75.3 / 91.7
& 55.8 / \textbf{80.0}
& 25.0 / 60.0
& \textbf{40.2} / 50.9
& 16.9 / 33.1
& 20.0 / 41.6
& 38.9 / 59.6
\\

& CalibRL
& 74.8 / 91.4
& 58.0 / 79.4
& 25.0 / 55.0
& 40.0 / 49.5
& 19.1 / 37.9
& 21.5 / \textbf{42.5}
& 39.7 / 59.3
\\

& \multicolumn{1}{>{\columncolor{light-gray}}l}{\textbf{Supervised}}
& \multicolumn{7}{l}{}
\\

& SFT
& 73.8 / 91.7
& 54.4 / 75.4
& 25.0 / 57.5
& 37.4 / 50.6
& 16.2 / 33.1
& 19.4 / 40.9
& 37.7 / 58.2
\\

& RL+SFT
& 74.1 / 91.0
& 53.6 / 75.4
& 22.5 / 60.0
& 38.4 / 49.7
& 17.3 / \textbf{38.2}
& 19.7 / 41.3
& 37.6 / 59.3
\\

& \model
& \textbf{76.1} / 91.7
& \textbf{59.0} / 79.0
& \textbf{30.0} / \textbf{67.5}
& 39.9 / 50.7
& \textbf{19.5} / 36.0
& \textbf{22.4} / 41.6
& \textbf{41.2} / \textbf{61.1}
\\

\bottomrule
\end{tabular}%
}
\caption{Main results on Qwen-2.5-Instruct with 0.5B and 1.5B backbones.
Each cell reports pass@1 / pass@8.}
\label{tab:reasoning}
\end{table*}

\begin{table}[!t]
\centering

\resizebox{0.9\linewidth}{!}{%
\begin{tabular}{clccc}
\toprule
&
\textbf{Model}
&
\textbf{Div-Eq}
&
\textbf{EAD}
&
\textbf{Emb-Div}
\\
\midrule

\multirow{12}{*}{\rotatebox{90}{\textbf{Qwen-2.5-0.5B}}}
& Base        & 0.5128 & 0.4802 & 0.0804 \\
& GRPO        & 0.4418 & 0.3880 & 0.0574 \\

\cmidrule(lr){2-5}

& Unlikely    & 0.3967 & 0.3169 & 0.1327 \\
& R1-Zero-Div & 0.9871 & 0.5938 & 0.2018 \\
& DQO         & 0.7215 & 0.6757 & 0.4566 \\
& DPH-RL      & 0.4708 & 0.4236 & 0.0664 \\

\cmidrule(lr){2-5}

& MixPolicy   & 0.4426 & 0.3978 & 0.0640 \\
& LUFFY       & 0.4695 & 0.4365 & 0.0706 \\
& CalibRL     & 0.4460 & 0.4039 & 0.0652 \\

\cmidrule(lr){2-5}

& SFT         & 0.4665 & 0.4198 & 0.0709 \\
& RL+SFT      & 0.4635 & 0.4209 & 0.0703 \\
& \model      & 0.4840 & 0.4490 & 0.0726 \\

\midrule
\midrule

\multirow{12}{*}{\rotatebox{90}{\textbf{Qwen-2.5-1.5B}}}
& Base        & 0.4824 & 0.4835 & 0.0757 \\
& GRPO        & 0.4014 & 0.3836 & 0.0549 \\

\cmidrule(lr){2-5}

& Unlikely    & 0.2859 & 0.4252 & 0.0977 \\
& R1-Zero-Div & 0.4927 & 0.4718 & 0.0743 \\
& DQO         & 0.5411 & 0.5566 & 0.2061 \\
& DPH-RL      & 0.4294 & 0.4159 & 0.0624 \\

\cmidrule(lr){2-5}

& MixPolicy   & 0.3933 & 0.3766 & 0.0570 \\
& LUFFY       & 0.4280 & 0.4179 & 0.0636 \\
& CalibRL     & 0.4007 & 0.3804 & 0.0542 \\

\cmidrule(lr){2-5}

& SFT         & 0.4365 & 0.4284 & 0.0694 \\
& RL+SFT      & 0.4390 & 0.4272 & 0.0689 \\
& \model      & 0.4348 & 0.4263 & 0.0639 \\

\bottomrule
\end{tabular}%
}
\caption{
Diversity results using pass@8 rollouts averaged over six benchmarks. Per-dataset results are provided in Appendix~\ref{app:exp}.
}
\label{tab:diversity}

\end{table}

\paragraph{Base models.}
We conduct experiments with two models from the Qwen2.5 family~\cite{qwen25}: Qwen2.5-0.5B-Instruct and Qwen2.5-1.5B-Instruct.
For all methods, we initialize post-training from the same checkpoint at each model scale.
The same checkpoint also serves as the base policy $\pi_{\mathrm{base}}$ used to construct the memory bank when applicable.

\paragraph{Datasets.}
We train all methods on the MATH training set~\cite{hendrycks2021math}, which contains 7,500 problems.
We evaluate the resulting models on six mathematical reasoning benchmarks spanning a broad range of difficulty levels and problem domains including MATH-500~\cite{lightman2024verify}, GSM8K~\cite{cobbe2021gsm8k}, OlympiadBench~\cite{he2024olympiadbench}, CollegeMath~\cite{tang2024mathscale}, AMC23, and Minerva-Math~\cite{lewkowycz2022minerva}.

\paragraph{Baselines.}
In addition to \textbf{GRPO}~\cite{shao2024deepseekmath}, we compare \model against two families of baselines.
(1) \emph{On-policy diversity methods}, which encourage exploration or diversity using only on-policy rollouts. 
These include \textbf{Unlikely}~\cite{he2025rewarding}, \textbf{R1-Zero-Div}~\cite{yao2025diversity}, \textbf{DQO}~\cite{chen2026posttraining}, and \textbf{DPH-RL}~\cite{li2026divergence}.
(2) \emph{Offline replay methods}, which additionally use the same memory-bank trajectories as \model. We further distinguish two ways of incorporating these trajectories. \emph{Mixed-policy replay} directly incorporates offline trajectories into the RL objective as off-policy samples, including \textbf{MixPolicy}, \textbf{LUFFY}~\cite{luffy}, and \textbf{CalibRL}~\cite{huang2026controllable}.
In contrast, \emph{supervised replay} uses the offline trajectories through a separate supervised objective rather than modifying the RL objective; we consider \textbf{SFT} and \textbf{RL+SFT} as representative controls. 
Further details of all baselines and implementation details are provided in Appendix~\ref{app:baselines}, \ref{app:impl}, respectively.

\paragraph{Evaluation.}
To evaluate reasoning-based abilities, we use metrics pass@1 and pass@8.
To evaluate response diversity, we compute three complementary metrics over the eight responses generated for each problem in the pass@8 evaluation, including 
\textbf{Equation diversity (Div-Eq)}~\cite{wu2025progress}, which measures diversity in mathematical derivations as the ratio of unique to total extracted LaTeX equations; 
\textbf{EAD}~\cite{liu2022rethinking}, which measures lexical diversity using expectation-adjusted Distinct-$n$, averaged over $n\in\{1,\ldots,5\}$; 
and \textbf{Embedding diversity (Emb-Div)}~\cite{reimers2019sentencebert}, which measures semantic diversity as one minus the mean pairwise cosine similarity between BGE-M3~\cite{chen2024bge} response embeddings.
Higher values indicate greater diversity for all three metrics.

\paragraph{Memory bank construction.}
For each training problem, we sample $N=128$ i.i.d. responses from the corresponding base policy and retain only the correct responses.
We encode the retained responses with BGE-M3~\cite{chen2024bge} using \texttt{[CLS]} pooling.
We then apply FPS under cosine distance and retain at most $k=3$ diverse trajectories per problem.
Problems for which the base policy produces no correct response are assigned no replay trajectories.
For a given base model, all methods that use offline replay share the same pre-constructed memory bank to ensure a fair comparison. The quality of the memory bank is evaluated in Appendix~\ref{app:bank}.

\subsection{Main Results}

Tables~\ref{tab:reasoning} and~\ref{tab:diversity} summarize reasoning performance and response diversity across six benchmarks.
\model achieves the highest average pass@1 and pass@8 at both model scales, while consistently mitigating the diversity reduction induced by standard GRPO.
Notably, On-Policy Diversity methods such as R1-Zero-Div and DQO can achieve much higher diversity scores, but often with weaker reasoning performance.
Meanwhile, naive supervised baselines (SFT and RL+SFT) achieve diversity comparable to \model, but yield lower reasoning accuracy, which shows continuous supervised replay may impose unnecessary constraints on policy update.
These findings suggest that neither explicitly maximizing diversity nor indiscriminately replaying correct trajectories necessarily achieves an effective balance between diversity preservation and policy improvement.
By selectively correcting suppressed base-model behaviors, \model mitigates diversity loss while allowing effective RL optimization.
Additional analysis restricted to correct responses is provided in Appendix~\ref{app:exp}.

\begin{figure}[!t]
    \centering
    \includegraphics[width=\linewidth]{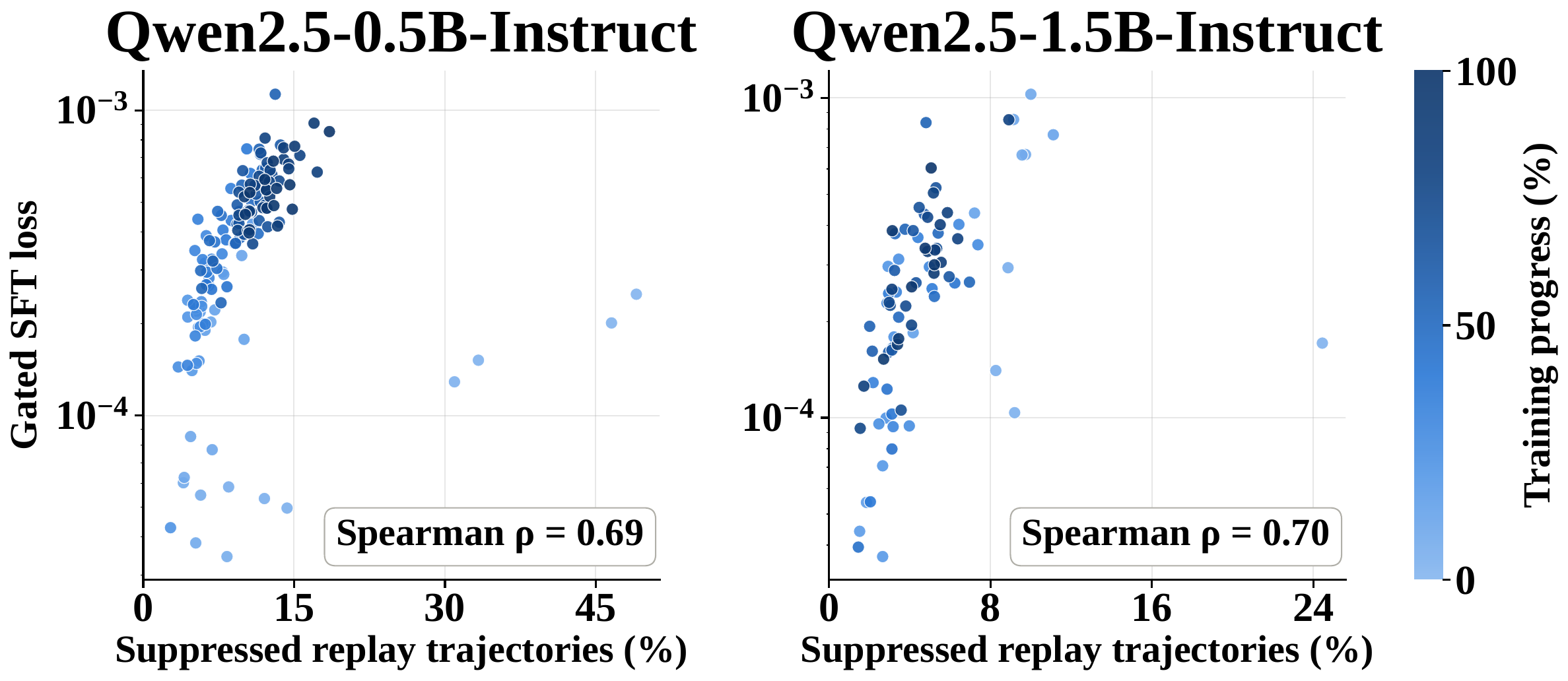}
\caption{
Relationship between trajectory suppression and gated replay loss during \model training.
Each point represents one training step; the x-axis shows the fraction of suppressed replay trajectories ($b(\tau)>0$), and the y-axis shows the gated replay loss.
}\label{fig:gate}
\end{figure}

\subsection{In-depth Analysis}

\paragraph{Does adaptive replay activate only when needed?}
Figure~\ref{fig:gate} examines whether \model's replay mechanism behaves as intended during training.
As more replay trajectories become suppressed, the gated replay loss generally increases.
Notably, several early-training steps exhibit a relatively high fraction of suppressed trajectories but little replay loss, because their suppression is still shallow and therefore receives little weight from the gate.
Overall, \model applies stronger correction only when suppression becomes sufficiently prevalent or pronounced, rather than continuously imposing an SFT objective.

\begin{figure}[!t]
    \centering
    \includegraphics[width=\linewidth]{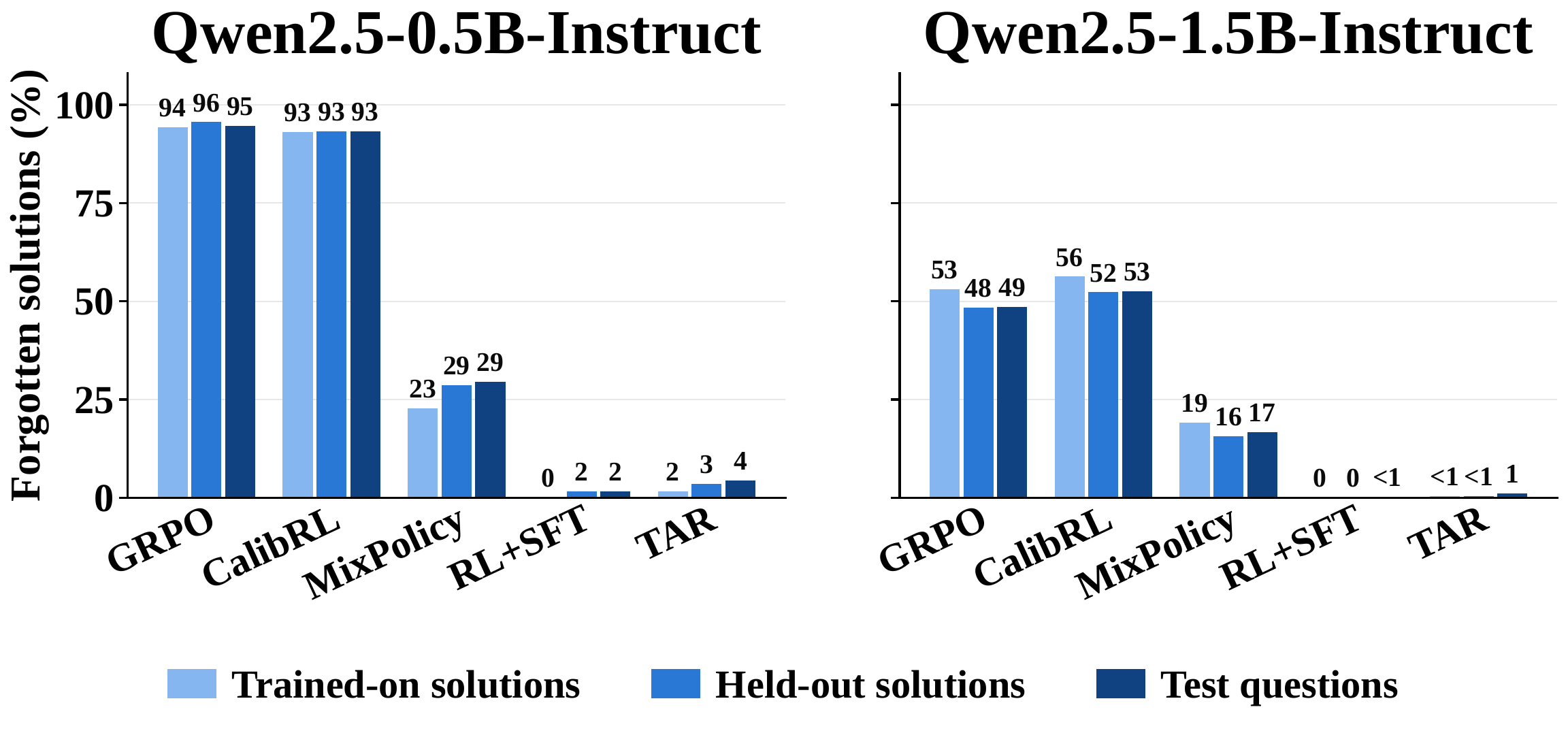}
\caption{Forgetting of base-model's correct solutions after training. A solution is counted as forgotten if the final policy assigns the complete solution at least $10\times$ lower sequence-level probability than the base model. We evaluate replay trajectories used during training (light blue), other correct base-model solutions to the same training questions (medium blue), and correct base-model solutions to unseen MATH-500 questions (dark blue).}
    \label{fig:retention}
\end{figure}

\begin{table}[t]
\centering

\resizebox{\columnwidth}{!}{%
\begin{tabular}{lcc}
\toprule
\textbf{Model}
&
\textbf{Reasoning}
&
\textbf{Diversity}
\\
\midrule

\model
& 41.2 / 61.1
& 0.4348 / 0.4263 / 0.0639
\\

\quad w/o token gate
& 40.4 / 59.0
& 0.4179 / 0.4004 / 0.0590
\\

\quad w/o rollout gate
& 38.5 / 58.7
& 0.4763 / 0.4755 / 0.0745
\\

\quad hard rollout gate
& 39.3 / 60.2
& 0.4714 / 0.4703 / 0.0731
\\

\bottomrule
\end{tabular}%
}
\caption{
Ablation study on Qwen-2.5-1.5B-Instruct, averaged over six benchmarks.
Reasoning reports pass@1 / pass@8; diversity reports Div-Eq / EAD / Emb-Div.
Per-dataset results are provided in Appendix~\ref{app:exp}.
}
\label{tab:ablation}

\end{table}

\paragraph{Does \model preserve base-policy solutions?}
Figure~\ref{fig:retention} probes whether correct solutions already present in the base policy remain likely after training.
GRPO and CalibRL exhibit severe forgetting, whereas MixPolicy partially alleviates this effect by incorporating off-policy trajectories.
In contrast, \model reduces forgetting to only $0$-$4\%$, achieving retention close to that of the naive RL+SFT.
Crucially, this improvement extends beyond replayed trajectories to held-out solutions and unseen test questions, suggesting that \model's corrective effect generalizes beyond the memory bank rather than merely preserving stored trajectories.

\subsection{Ablation Study}
\label{sec:ablation}

\begin{figure*}[!t]
    \centering
    \includegraphics[width=0.78\linewidth]{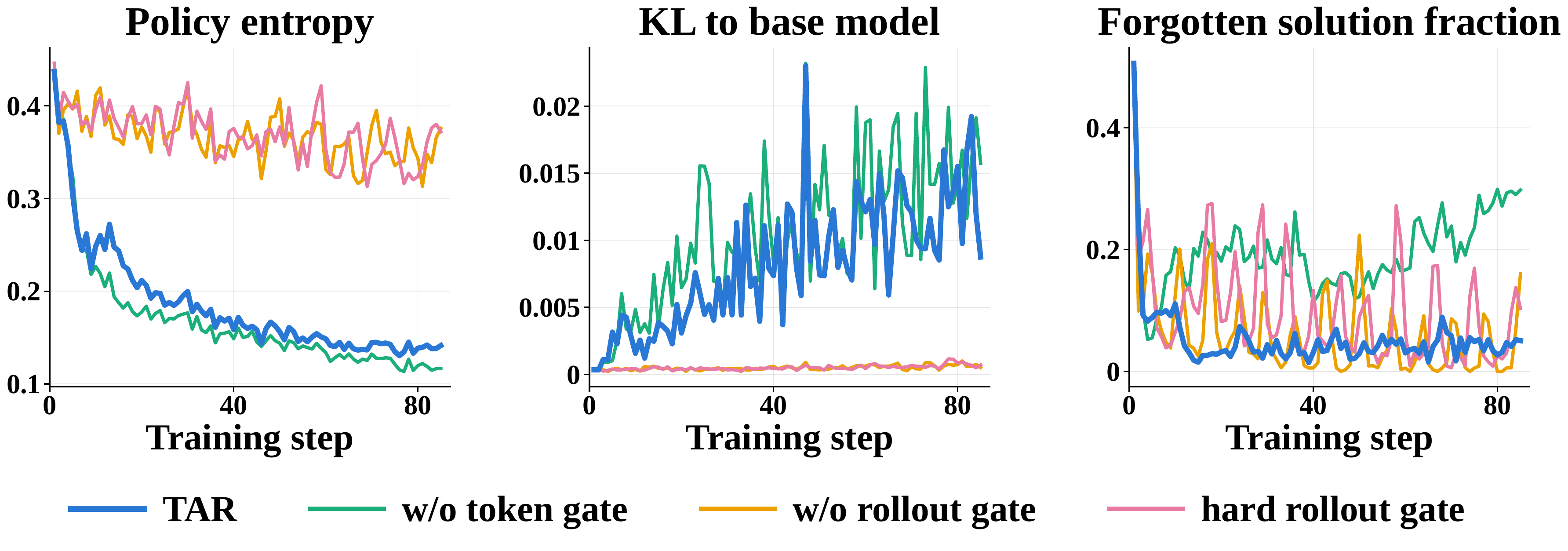}
\caption{Training dynamics for \model and different ablated variants.}
    \label{fig:ablation-dynamics}
\end{figure*}

We ablate the rollout-level gate $g(\tau)$ and token-level gate $w_t$ in \model's adaptive replay mechanism.
Specifically, ``w/o token gate'' replaces the token weights with uniform weights; ``w/o rollout gate'' sets $g(\tau)=1$ for all replay trajectories; and ``hard rollout gate'' uses $g(\tau)=\mathbbm{1}[b(\tau)>0]$.
Table~\ref{tab:ablation} reports the final performance, while Figure~\ref{fig:ablation-dynamics} shows the training dynamics.

\paragraph{The soft rollout gate enables effective policy adaptation.} 
Removing the rollout gate or replacing it with a hard binary gate places the policy in a qualitatively different training regime. As shown in Figure~\ref{fig:ablation-dynamics}, both variants remain close to the base model throughout training, with near-zero KL divergence and only modest entropy reduction.
Thus, their apparently low forgetting in Figure~\ref{fig:ablation-dynamics} and high diversity in Table~\ref{tab:ablation} largely reflect limited policy adaptation rather than more effective preservation.
These results show that the soft rollout gate is important for balancing corrective replay with meaningful policy improvement.

\paragraph{The token gate is necessary to localize correction.}
In contrast, ``w/o token gate'' exhibits entropy and KL dynamics similar to \model, indicating that policy learning proceeds normally.
However, uniformly distributing the same trajectory-level corrective signal across all tokens substantially increases forgetting.
Consistently, removing the token gate reduces both reasoning performance and all three diversity metrics.

\section{Related Work}

\paragraph{Diversity in RLVR.}
Recent work has shown that RLVR can sharpen the output distribution and reduce reasoning diversity, even when reasoning performance improves~\cite{yue2025does,wu2025progress,he2025rewarding}.
To mitigate this effect, existing methods encourage broader exploration by modifying the learning signal over sampled responses, e.g., through likelihood-based reweighting, entropy regularization, set-level diversity rewards, or mass-covering objectives~\cite{he2025rewarding,yao2025diversity,chen2026posttraining,li2026divergence}.
Our work addresses a complementary failure mode. Once a valid reasoning mode becomes sufficiently under-sampled, it may no longer appear in the current rollout group, in which case reweighting the observed responses cannot directly provide a corrective signal for that missing mode. \model instead preserves diverse correct modes from the base policy and can reinforce them independently of whether they are resampled on-policy.

\paragraph{Off-policy guidance and replay.}
Several approaches incorporate offline or expert trajectories into RL post-training, including methods that mix demonstrations with on-policy rollouts, use expert trajectories to calibrate policy updates, or replay past behaviors to mitigate forgetting~\cite{luffy,huang2026controllable,yuan2025more}.
\model differs in both how replay trajectories are obtained and how they are used.
Rather than relying on external demonstrations, \model stores diverse correct solutions already present in the base policy.
More importantly, these trajectories are not continuously imitated: replay is activated only when a stored behavior has been suppressed relative to the base policy, with the corrective signal further localized to the suppressed token transitions.

\section{Conclusion}

Our study highlights a mismatch between the RLVR objective and its finite-sample optimization dynamics: finite on-policy sampling can progressively suppress equally rewarded reasoning modes, which rarely recover once sufficiently under-sampled. \model alleviates this issue by selectively restoring diverse correct behaviors inherited from the base policy. Our findings further show that diversity preservation requires more than simply retaining the original policy distribution: indiscriminate replay can restrict policy adaptation, whereas targeted correction preserves alternative reasoning behaviors while still allowing meaningful policy improvement. Across six reasoning benchmarks and two model scales, \model achieves a better balance between reasoning performance and diversity preservation than competing post-training approaches.

\section*{Limitations}

First, due to computational constraints, our experiments are limited to relatively small language models and mathematical reasoning benchmarks. Whether these findings extend to larger models and other reasoning domains remains to be investigated. Second, \model relies on a finite memory bank constructed from correct base-model trajectories. Its ability to preserve reasoning diversity is therefore constrained by the reasoning behaviors available in the base policy and the coverage of the sampled memory bank. In particular, it cannot directly restore reasoning modes absent from the memory bank. Finally, memory-bank construction and adaptive replay introduce additional computational overhead from offline sampling, embedding, and replay updates. We leave a systematic investigation of the trade-off between computational efficiency and diversity preservation to future work.

\bibliography{sample-base}

@article{guo2025deepseek,
  title={{DeepSeek-R1} incentivizes reasoning in {LLMs} through reinforcement learning},
  author={
    Guo, Daya and
    Yang, Dejian and
    Zhang, Haowei and
    Song, Junxiao and
    Wang, Peiyi and
    Zhu, Qihao and
    Xu, Runxin and
    Zhang, Ruoyu and
    Ma, Shirong and
    Bi, Xiao and
    Zhang, Xiaokang and
    Yu, Xingkai and
    Wu, Yu and
    Wu, Z. F. and
    Gou, Zhibin and
    Shao, Zhihong and
    Li, Zhuoshu and
    Gao, Ziyi and
    Liu, Aixin and
    Xue, Bing and
    Wang, Bingxuan and
    Wu, Bochao and
    Feng, Bei and
    Lu, Chengda and
    Zhao, Chenggang and
    Deng, Chengqi and
    Ruan, Chong and
    Dai, Damai and
    Chen, Deli and
    Ji, Dongjie and
    Li, Erhang and
    Lin, Fangyun and
    Dai, Fucong and
    Luo, Fuli and
    Hao, Guangbo and
    Chen, Guanting and
    Li, Guowei and
    Zhang, H. and
    Xu, Hanwei and
    Ding, Honghui and
    Gao, Huazuo and
    Qu, Hui and
    Li, Hui and
    Guo, Jianzhong and
    Li, Jiashi and
    Chen, Jingchang and
    Yuan, Jingyang and
    Tu, Jinhao and
    Qiu, Junjie and
    Li, Junlong and
    Cai, J. L. and
    Ni, Jiaqi and
    Liang, Jian and
    Chen, Jin and
    Dong, Kai and
    Hu, Kai and
    You, Kaichao and
    Gao, Kaige and
    Guan, Kang and
    Huang, Kexin and
    Yu, Kuai and
    Wang, Lean and
    Zhang, Lecong and
    Zhao, Liang and
    Wang, Litong and
    Zhang, Liyue and
    Xu, Lei and
    Xia, Leyi and
    Zhang, Mingchuan and
    Zhang, Minghua and
    Tang, Minghui and
    Zhou, Mingxu and
    Li, Meng and
    Wang, Miaojun and
    Li, Mingming and
    Tian, Ning and
    Huang, Panpan and
    Zhang, Peng and
    Wang, Qiancheng and
    Chen, Qinyu and
    Du, Qiushi and
    Ge, Ruiqi and
    Zhang, Ruisong and
    Pan, Ruizhe and
    Wang, Runji and
    Chen, R. J. and
    Jin, R. L. and
    Chen, Ruyi and
    Lu, Shanghao and
    Zhou, Shangyan and
    Chen, Shanhuang and
    Ye, Shengfeng and
    Wang, Shiyu and
    Yu, Shuiping and
    Zhou, Shunfeng and
    Pan, Shuting and
    Li, S. S. and
    Zhou, Shuang and
    Wu, Shaoqing and
    Yun, Tao and
    Pei, Tian and
    Sun, Tianyu and
    Wang, T. and
    Zeng, Wangding and
    Liu, Wen and
    Liang, Wenfeng and
    Gao, Wenjun and
    Yu, Wenqin and
    Zhang, Wentao and
    Xiao, W. L. and
    An, Wei and
    Liu, Xiaodong and
    Wang, Xiaohan and
    Chen, Xiaokang and
    Nie, Xiaotao and
    Cheng, Xin and
    Liu, Xin and
    Xie, Xin and
    Liu, Xingchao and
    Yang, Xinyu and
    Li, Xinyuan and
    Su, Xuecheng and
    Lin, Xuheng and
    Li, X. Q. and
    Jin, Xiangyue and
    Shen, Xiaojin and
    Chen, Xiaosha and
    Sun, Xiaowen and
    Wang, Xiaoxiang and
    Song, Xinnan and
    Zhou, Xinyi and
    Wang, Xianzu and
    Shan, Xinxia and
    Li, Y. K. and
    Wang, Y. Q. and
    Wei, Y. X. and
    Zhang, Yang and
    Xu, Yanhong and
    Li, Yao and
    Zhao, Yao and
    Sun, Yaofeng and
    Wang, Yaohui and
    Yu, Yi and
    Zhang, Yichao and
    Shi, Yifan and
    Xiong, Yiliang and
    He, Ying and
    Piao, Yishi and
    Wang, Yisong and
    Tan, Yixuan and
    Ma, Yiyang and
    Liu, Yiyuan and
    Guo, Yongqiang and
    Ou, Yuan and
    Wang, Yuduan and
    Gong, Yue and
    Zou, Yuheng and
    He, Yujia and
    Xiong, Yunfan and
    Luo, Yuxiang and
    You, Yuxiang and
    Liu, Yuxuan and
    Zhou, Yuyang and
    Zhu, Y. X. and
    Huang, Yanping and
    Li, Yaohui and
    Zheng, Yi and
    Zhu, Yuchen and
    Ma, Yunxian and
    Tang, Ying and
    Zha, Yukun and
    Yan, Yuting and
    Ren, Z. Z. and
    Ren, Zehui and
    Sha, Zhangli and
    Fu, Zhe and
    Xu, Zhean and
    Xie, Zhenda and
    Zhang, Zhengyan and
    Hao, Zhewen and
    Ma, Zhicheng and
    Yan, Zhigang and
    Wu, Zhiyu and
    Gu, Zihui and
    Zhu, Zijia and
    Liu, Zijun and
    Li, Zilin and
    Xie, Ziwei and
    Song, Ziyang and
    Pan, Zizheng and
    Huang, Zhen and
    Xu, Zhipeng and
    Zhang, Zhongyu and
    Zhang, Zhen
  },
  journal={Nature},
  volume={645},
  number={8081},
  pages={633--638},
  year={2025}
}

@article{korbak2022reinforcement,
  title={On reinforcement learning and distribution matching for fine-tuning language models with no catastrophic forgetting},
  author={Korbak, Tomasz and Elsahar, Hady and Kruszewski, Germ{\'a}n and Dymetman, Marc},
  journal={Advances in Neural Information Processing Systems},
  volume={35},
  pages={16203--16220},
  year={2022}
}

@inproceedings{yue2025does,
  title={Does Reinforcement Learning Really Incentivize Reasoning Capacity in {LLMs} Beyond the Base Model?},
  author={Yue, Yang and Chen, Zhiqi and Lu, Rui and Zhao, Andrew and Wang, Zhaokai and Yue, Yang and Song, Shiji and Huang, Gao},
  booktitle={Advances in Neural Information Processing Systems},
  volume={38},
  year={2025}
}

@inproceedings{zhang2025noveltybench,
  title={{NoveltyBench}: Evaluating Language Models for Human-like Diversity},
  author={Zhang, Yiming and Diddee, Harshita and Holm, Susan and Liu, Hanchen and Liu, Xinyue and Samuel, Vinay and Wang, Barry and Ippolito, Daphne},
  booktitle={Second Conference on Language Modeling},
  year={2025}
}

@article{jaech2024openai,
  title={Openai o1 system card},
  author={Jaech, Aaron and Kalai, Adam and Lerer, Adam and Richardson, Adam and El-Kishky, Ahmed and Low, Aiden and Helyar, Alec and Madry, Aleksander and Beutel, Alex and Carney, Alex and others},
  journal={arXiv preprint arXiv:2412.16720},
  year={2024}
}

@article{yang2025qwen3,
  title={Qwen3 technical report},
  author={Yang, An and Li, Anfeng and Yang, Baosong and Zhang, Beichen and Hui, Binyuan and Zheng, Bo and Yu, Bowen and Gao, Chang and Huang, Chengen and Lv, Chenxu and others},
  journal={arXiv preprint arXiv:2505.09388},
  year={2025}
}

@article{wu2025invisible,
  title={The invisible leash: Why rlvr may or may not escape its origin},
  author={Wu, Fang and Xuan, Weihao and Lu, Ximing and Liu, Mingjie and Dong, Yi and Harchaoui, Zaid and Choi, Yejin},
  journal={arXiv preprint arXiv:2507.14843},
  year={2025}
}

@inproceedings{chen2025retaining,
  title={Retaining by Doing: The Role of On-Policy Data in Mitigating Forgetting},
  author={Chen, Howard and Razin, Noam and Narasimhan, Karthik R. and Chen, Danqi},
  booktitle={Proceedings of the 43rd International Conference on Machine Learning},
  series={Proceedings of Machine Learning Research},
  volume={306},
  pages={16479--16494},
  year={2026}
}

@inproceedings{gx2025kl,
  title={KL-Regularized Reinforcement Learning for Generative Modelling is Designed to Mode Collapse},
  author={GX-Chen, Anthony and Prakash, Jatin and Guo, Jeff and Fergus, Rob and Ranganath, Rajesh},
  booktitle={The Fourteenth International Conference on Learning Representations},
  year={2026}
}

@inproceedings{he2025rewarding,
  title={Rewarding the unlikely: Lifting grpo beyond distribution sharpening},
  author={He, Andre Wang and Fried, Daniel and Welleck, Sean},
  booktitle={Proceedings of the 2025 Conference on Empirical Methods in Natural Language Processing},
  pages={25548--25560},
  year={2025}
}

@article{shao2024deepseekmath,
  title={Deepseekmath: Pushing the limits of mathematical reasoning in open language models},
  author={Shao, Zhihong and Wang, Peiyi and Zhu, Qihao and Xu, Runxin and Song, Junxiao and Bi, Xiao and Zhang, Haowei and Zhang, Mingchuan and Li, YK and Wu, Yang and Guo, Daya},
  journal={arXiv preprint arXiv:2402.03300},
  year={2024}
}

@inproceedings{vllm,
  title={Efficient Memory Management for Large Language Model Serving with PagedAttention},
  author={Woosuk Kwon and Zhuohan Li and Siyuan Zhuang and Ying Sheng and Lianmin Zheng and Cody Hao Yu and Joseph E. Gonzalez and Hao Zhang and Ion Stoica},
  booktitle={Proceedings of the ACM SIGOPS 29th Symposium on Operating Systems Principles},
  year={2023}
}

@article{yuan2025more,
  title={More Than One Teacher: Adaptive Multi-Guidance Policy Optimization for Diverse Exploration},
  author={Yuan, Xiaoyang and Ding, Yujuan and Bin, Yi and Shao, Wenqi and Cai, Jinyu and Song, Jingkuan and Yang, Yang and Shen, Heng Tao},
  journal={arXiv preprint arXiv:2510.02227},
  year={2025}
}

@inproceedings{zhou2025expo,
  title={{ExPO}: Unlocking Hard Reasoning with Self-Explanation-Guided Reinforcement Learning},
  author={Zhou, Ruiyang and Li, Shuozhe and Zhang, Amy and Liu Leqi},
  booktitle={Advances in Neural Information Processing Systems},
  volume={38},
  year={2025}
}

@article{dong2023raft,
  title={{RAFT}: Reward rAnked FineTuning for Generative Foundation Model Alignment},
  author={Dong, Hanze and Xiong, Wei and Goyal, Deepanshu and Zhang, Yihan and Chow, Winnie and Pan, Rui and Diao, Shizhe and Zhang, Jipeng and Shum, Kashun and Zhang, Tong},
  journal={Transactions on Machine Learning Research},
  year={2023}
}

@inproceedings{wang2025beyond,
  title={Beyond the 80/20 Rule: High-Entropy Minority Tokens Drive Effective Reinforcement Learning for {LLM} Reasoning},
  author={Wang, Shenzhi and Yu, Le and Gao, Chang and Zheng, Chujie and Liu, Shixuan and Lu, Rui and Dang, Kai and Chen, Xionghui and Yang, Jianxin and Zhang, Zhenru and Liu, Yuqiong and Yang, An and Zhao, Andrew and Yue, Yang and Song, Shiji and Yu, Bowen and Huang, Gao and Lin, Junyang},
  booktitle={Advances in Neural Information Processing Systems},
  volume={38},
  year={2025}
}

@inproceedings{cheng2026reasoning,
  title={Reasoning with exploration: An entropy perspective},
  author={Cheng, Daixuan and Huang, Shaohan and Zhu, Xuekai and Dai, Bo and Zhao, Xin and Zhang, Zhenliang and Wei, Furu},
  booktitle={Proceedings of the AAAI Conference on Artificial Intelligence},
  volume={40},
  number={36},
  pages={30377--30385},
  year={2026}
}

@article{ppo,
  title={Proximal policy optimization algorithms},
  author={Schulman, John and Wolski, Filip and Dhariwal, Prafulla and Radford, Alec and Klimov, Oleg},
  journal={arXiv preprint arXiv:1707.06347},
  year={2017}
}

@article{fps,
  title={Clustering to minimize the maximum intercluster distance},
  author={Gonzalez, Teofilo F},
  journal={Theoretical computer science},
  volume={38},
  pages={293--306},
  year={1985},
  publisher={Elsevier}
}

@article{qwen25,
      title={Qwen2.5 Technical Report}, 
      author={{Qwen Team}},
      year={2025},
      journal={arXiv preprint arXiv:2412.15115},
}

@inproceedings{hendrycks2021math,
  title={Measuring Mathematical Problem Solving With the MATH Dataset},
  author={Hendrycks, Dan and Burns, Collin and Kadavath, Saurav and Arora, Akul and
          Basart, Steven and Tang, Eric and Song, Dawn and Steinhardt, Jacob},
  booktitle={Advances in Neural Information Processing Systems},
  year={2021}
}

@inproceedings{lightman2024verify,
  title={Let's Verify Step by Step},
  author={Lightman, Hunter and Kosaraju, Vineet and Burda, Yura and
          Edwards, Harri and Baker, Bowen and Lee, Teddy and Leike, Jan and
          Schulman, John and Sutskever, Ilya and Cobbe, Karl},
  booktitle={International Conference on Learning Representations},
  year={2024}
}

@article{cobbe2021gsm8k,
  title={Training Verifiers to Solve Math Word Problems},
  author={Cobbe, Karl and Kosaraju, Vineet and Bavarian, Mohammad and
          Chen, Mark and Jun, Heewoo and Kaiser, Lukasz and Plappert, Matthias and
          Tworek, Jerry and Hilton, Jacob and Nakano, Reiichiro and
          Hesse, Christopher and Schulman, John},
  journal={arXiv preprint arXiv:2110.14168},
  year={2021}
}

@inproceedings{he2024olympiadbench,
  title={OlympiadBench: A Challenging Benchmark for Promoting AGI with
         Olympiad-Level Bilingual Multimodal Scientific Problems},
  author={He, Chaoqun and Luo, Renjie and Bai, Yuzhuo and Hu, Shengding and
          Thai, Zhen and Shen, Junhao and Hu, Jinyi and Han, Xu and
          Huang, Yujie and Zhang, Yuxiang and Liu, Jie and Qi, Lei and
          Liu, Zhiyuan and Sun, Maosong},
  booktitle={Proceedings of the 62nd Annual Meeting of the Association for
             Computational Linguistics},
  year={2024}
}

@inproceedings{tang2024mathscale,
  title={MathScale: Scaling Instruction Tuning for Mathematical Reasoning},
  author={Tang, Zhengyang and Zhang, Xingxing and Wang, Benyou and Wei, Furu},
  booktitle={Proceedings of the 41st International Conference on Machine Learning},
  year={2024}
}

@article{lewkowycz2022minerva,
  title={Solving Quantitative Reasoning Problems with Language Models},
  author={Lewkowycz, Aitor and Andreassen, Anders and Dohan, David and
          Dyer, Ethan and Michalewski, Henryk and Ramasesh, Vinay and
          Slone, Ambrose and Anil, Cem and Schlag, Imanol and
          Gutman-Solo, Theo and Wu, Yuhuai and Neyshabur, Behnam and
          Gur-Ari, Guy and Misra, Vedant},
  journal={Advances in Neural Information Processing Systems},
  year={2022}
}

@inproceedings{yao2025diversity,
  title={Diversity-Aware Policy Optimization for Large Language Model Reasoning},
  author={Yao, Jian and Cheng, Ran and Wu, Xingyu and Wu, Jibin and Tan, Kay Chen},
  booktitle={Advances in Neural Information Processing Systems},
  year={2025}
}

@inproceedings{chen2026posttraining,
  title={Post-training Large Language Models for Diverse High-Quality Responses},
  author={Chen, Yilei and Chakraborty, Souradip and Wolf, Lorenz and
          Paschalidis, Ioannis and Pacchiano, Aldo},
  booktitle={International Conference on Learning Representations},
  year={2026}
}

@inproceedings{li2026divergence,
  title={The Choice of Divergence: A Neglected Key to Mitigating Diversity Collapse in Reinforcement Learning with Verifiable Reward},
  author={Li, Long and Zhou, Zhijian and Hao, Jiaran and Liu, Jason and Miao, Yanting and Pang, Wei and Tan, Xiaoyu and Chu, Wei and Wang, Zhe and Pan, Shirui and Qu, Chao and Qi, Yuan},
  booktitle={International Conference on Learning Representations},
  year={2026}
}

@inproceedings{luffy,
  title={Learning to Reason under Off-Policy Guidance},
  author={Yan, Jianhao and Li, Yafu and Hu, Zican and Wang, Zhi and
          Cui, Ganqu and Qu, Xiaoye and Cheng, Yu and Zhang, Yue},
  booktitle={Advances in Neural Information Processing Systems},
  year={2025}
}

@inproceedings{huang2026controllable,
  title={Controllable Exploration in Hybrid-Policy RLVR for Multi-Modal Reasoning},
  author={Huang, Zhuoxu and Jia, Mengxi and Sun, Hao and Li, Xuelong and Han, Jungong},
  booktitle={International Conference on Learning Representations},
  year={2026}
}

@inproceedings{wu2025progress,
  title={Progress or Regress? Self-Improvement Reversal in Post-Training},
  author={Wu, Ting and Li, Xuefeng and Liu, Pengfei},
  booktitle={International Conference on Learning Representations},
  year={2025}
}

@inproceedings{liu2022rethinking,
  title={Rethinking and Refining the Distinct Metric},
  author={Liu, Siyang and Sabour, Sahand and Zheng, Yinhe and Ke, Pei and
          Zhu, Xiaoyan and Huang, Minlie},
  booktitle={Proceedings of the 60th Annual Meeting of the Association for
             Computational Linguistics (Volume 2: Short Papers)},
  pages={762--770},
  year={2022}
}

@inproceedings{chen2024bge,
  title={{M}3-Embedding: Multi-Linguality, Multi-Functionality, Multi-Granularity Text Embeddings Through Self-Knowledge Distillation},
  author={Chen, Jianlyu and Xiao, Shitao and Zhang, Peitian and Luo, Kun and Lian, Defu and Liu, Zheng},
  booktitle={Findings of the Association for Computational Linguistics: ACL 2024},
  pages={2318--2335},
  year={2024}
}

@inproceedings{reimers2019sentencebert,
  title={Sentence-BERT: Sentence Embeddings using Siamese BERT-Networks},
  author={Reimers, Nils and Gurevych, Iryna},
  booktitle={Proceedings of the 2019 Conference on Empirical Methods in Natural Language Processing
             and the 9th International Joint Conference on Natural Language Processing},
  year={2019}
}

@inproceedings{sheng2024hybridflow,
  title={{HybridFlow}: A Flexible and Efficient {RLHF} Framework},
  author={Sheng, Guangming and Zhang, Chi and Ye, Zilingfeng and Wu, Xibin and Zhang, Wang and Zhang, Ru and Peng, Yanghua and Lin, Haibin and Wu, Chuan},
  booktitle={Proceedings of the Twentieth European Conference on Computer Systems},
  pages={1279--1297},
  year={2025}
}

\appendix

\section{Baselines in Detail}
\label{app:baselines}

\paragraph{On-policy diversity methods.}
These baselines promote exploration or diversity without using our memory bank.
\begin{itemize}[leftmargin=*,nosep]
    \item \textbf{Unlikely}~\cite{he2025rewarding} mitigates GRPO's preference for high-likelihood solutions by down-weighting rewards according to the likelihood rank of each response within the rollout group.
    Consequently, lower-likelihood correct responses receive relatively larger updates.

    \item \textbf{R1-Zero-Div}~\cite{yao2025diversity} adds an entropy-based diversity regularizer on responses with positive advantage, encouraging exploration among correct solutions while leaving incorrect responses unaffected.

    \item \textbf{DQO}~\cite{chen2026posttraining} measures diversity among the responses in each rollout group using a determinantal point process.
    Each response receives an additional reward based on its marginal contribution to the diversity of the group.

    \item \textbf{DPH-RL}~\cite{li2026divergence} separates training problems according to base-model performance.
    Standard RL is applied to the exploration subset, while responses pre-sampled from the base model are rehearsed on problems the base model already solves reliably using a mass-covering divergence.
    We follow the reference-free configuration of the original method and disable the reverse-KL penalty on rollout samples.
\end{itemize}

\paragraph{Offline replay methods.}
These baselines use the same memory bank as \model but differ in how the stored trajectories are incorporated into training.
\begin{itemize}[leftmargin=*,nosep]
    \item \textbf{Mix-Policy} directly appends the stored trajectories of each prompt to its GRPO rollout group and computes advantages over the combined on-policy and replay samples.
    \item \textbf{LUFFY}~\cite{luffy} similarly mixes off-policy demonstrations with on-policy rollouts, but uses policy shaping rather than standard importance weighting to amplify gradients on low-probability demonstration tokens.
    \item \textbf{CalibRL}~\cite{huang2026controllable} uses demonstration trajectories as expert references to calibrate the learning signal on on-policy responses through an auxiliary objective.
    Since the original method assumes a single expert trajectory, we compute the calibration term separately for each stored trajectory and average across the available trajectories for each prompt.
    \item \textbf{SFT} fine-tunes the base model solely on the stored trajectories using the standard token-level negative log-likelihood objective, without RL rollouts or rewards.
    \item \textbf{RL+SFT} combines the standard GRPO objective with an ungated SFT loss over all stored trajectories. It therefore provides a direct control for evaluating the benefit of selective adaptive replay in \model.
\end{itemize}

\section{Implementation Details}
\label{app:impl}

\subsection{Training}

\paragraph{Shared hyperparameters.}
Table~\ref{tab:hparams-shared} lists the configuration shared by all RL methods.

\paragraph{Baseline hyperparameters.}
Unless noted otherwise, we follow the recommended settings of each original paper or codebase.
All settings shared with GRPO are kept identical to Table~\ref{tab:hparams-shared}.
Table~\ref{tab:hparams-baselines} summarizes the method-specific hyperparameters.

\paragraph{\model hyperparameters.}
We set the replay loss coefficient $\lambda_{\mathrm{rep}}$ to 1.
It is worth noting that $\pi_{\text{base}}$ in Eq.~\eqref{eq:delta} is the frozen reference model, which is already maintained for the KL term, so the gate requires no additional model.

For both model scales, each training run completes within 15 hours of wall-clock time on the GPU configurations listed in Table~\ref{tab:hparams-shared}, corresponding to at most 30 GPU-hours per run.

\subsection{Evaluation}

For all methods on all six benchmarks, we use temperature 0.6 and top-$p$ 0.95 for evaluation.
We compute pass@1 using one sampled response per problem and pass@8 using eight sampled responses, where a problem is considered solved if at least one response is correct.
We report the unweighted average across the six benchmarks.

\subsection{Artifacts and Licenses}

Our implementation uses verl~\cite{sheng2024hybridflow} for training and vLLM~\cite{vllm} for rollout generation, both under Apache 2.0. We fine-tune Qwen2.5-0.5B-Instruct and Qwen2.5-1.5B-Instruct~\cite{qwen25} (Apache 2.0) and use BGE-M3~\cite{chen2024bge} (MIT) for memory-bank embeddings. We plan to release our code under Apache 2.0, retaining applicable copyright and license notices and documenting modifications to upstream code.

\begin{table}[t]
\centering

\resizebox{\linewidth}{!}{%
\begin{tabular}{lcc}
\toprule
\textbf{Hyperparameter} & \textbf{0.5B} & \textbf{1.5B} \\
\midrule
Training epochs                     & 5 & 3 \\
GPUs                                & 2$\times$A6000 & 2$\times$A100 \\
\midrule
\rowcolor{light-gray}\multicolumn{3}{c}{\textbf{Shared across scales}}\\
Train batch size & \multicolumn{2}{c}{256} \\
PPO mini batch size & \multicolumn{2}{c}{128} \\
PPO micro batch size per GPU & \multicolumn{2}{c}{8} \\
GRPO group size                     & \multicolumn{2}{c}{8} \\
Optimizer                           & \multicolumn{2}{c}{AdamW} \\
Learning rate                       & \multicolumn{2}{c}{$1\times10^{-6}$} \\
Weight decay                        & \multicolumn{2}{c}{0.01} \\
Gradient clipping                   & \multicolumn{2}{c}{1.0} \\
PPO clipping ratio                  & \multicolumn{2}{c}{0.2} \\
KL loss coefficient $\beta$         & \multicolumn{2}{c}{0.001} \\
Entropy coefficient                 & \multicolumn{2}{c}{0.0} \\
Maximum prompt length               & \multicolumn{2}{c}{1024} \\
Maximum response length             & \multicolumn{2}{c}{1024} \\
Rollout temperature                 & \multicolumn{2}{c}{1.0} \\
Rollout top-$p$                     & \multicolumn{2}{c}{1.0} \\
\bottomrule
\end{tabular}
}
\caption{Shared training configuration.}
\label{tab:hparams-shared}
\end{table}

\begin{table}[t]
\centering
\small
\resizebox{\linewidth}{!}{%
\begin{tabular}{p{0.22\linewidth}p{0.70\linewidth}}
\toprule
\textbf{Method} & \textbf{Method-specific settings} \\
\midrule
Unlikely & Likelihood-rank penalty $\beta=0.25$ \\
R1-Zero-Div & Entropy regularization weight $\omega=0.01$; applied only to responses with $\hat A>0$ \\
DQO & Diversity reward weight $\alpha=1.0$; encoder: all-MiniLM-L6-v2~\cite{reimers2019sentencebert}; \\
DPH-RL & JS-divergence weight $0.0125$ \\
LUFFY & Policy-shaping parameter $\gamma=0.1$ \\
CalibRL & Expert-guidance coefficient $c=0.1$; LeakyReLU negative slope $0.5$ \\
\model & Replay loss coefficient $\lambda_{\mathrm{rep}}=1$ \\
\bottomrule
\end{tabular}
}
\caption{Method-specific hyperparameters.}
\label{tab:hparams-baselines}
\end{table}

\section{More Discussion on Memory Bank}\label{app:bank}

\begin{table}[t]
\centering
\small
\resizebox{\linewidth}{!}{%
\begin{tabular}{lcc}
\toprule
 & \textbf{0.5B} & \textbf{1.5B} \\
\midrule
Problems covered      & 5{,}978 & 6{,}554 \\
\quad (of 7{,}500)    & 79.7\% & 87.4\% \\
Problems with 3 / 2 / 1 traj. & 5363/226/389 & 6186/139/229 \\
Stored trajectories   & 16{,}930 & 19{,}065 \\
\bottomrule
\end{tabular}
}
\caption{Memory bank statistics ($N=128$ samples per problem, FPS with $k=3$).}
\label{tab:membank}
\end{table}

\begin{figure}[!t]
    \centering
    \includegraphics[width=\linewidth]{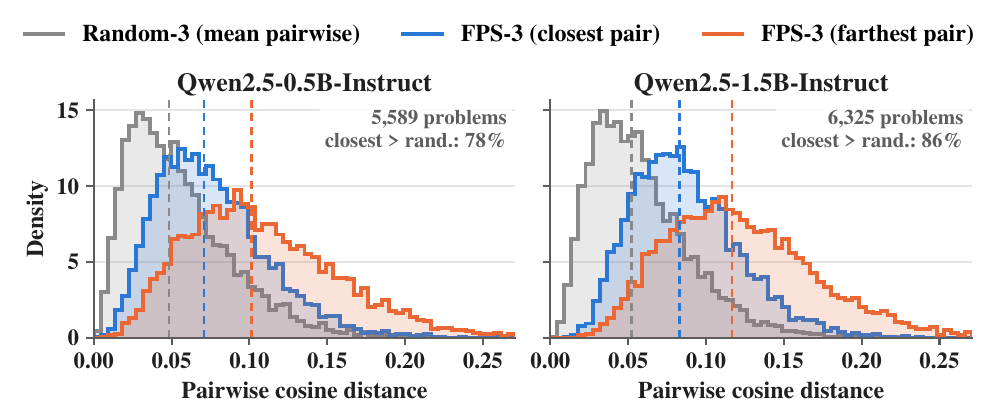}
\caption{Pairwise distance distributions of FPS-selected and randomly selected correct trajectories. Gray shows the mean pairwise cosine distance of three randomly selected correct trajectories, while blue and orange show the closest and farthest pair among $k=3$ FPS-selected trajectories. Dashed lines indicate medians.}\label{fig:memorybank}
\end{figure}

For each training question, we draw $N=128$ samples from the base policy, with temperature $0.6$ and top-$p=0.95$.
We score them with the training reward function and keep the correct ones.
Each correct response is embedded with BGE-M3~\cite{chen2024bge}, using \texttt{[CLS]} pooling and truncation at 1,024 tokens.
We then apply FPS under cosine distance to select at most $k=3$ diverse trajectories per problem.
Starting from a randomly selected correct trajectory, FPS iteratively selects the candidate that maximizes its minimum cosine distance to the already selected trajectories.
A separate bank is built for every base model.
Table~\ref{tab:membank} reports the bank statistics.

We further evaluate whether FPS promotes semantic diversity in the memory bank by comparing it with random selection of the same size.
For each problem with $M\geq 2$ correct samples, both methods select $\min(k,M)$ trajectories.
We compare the minimum and maximum pairwise cosine distances among FPS-selected trajectories with the mean pairwise distance among trajectories selected uniformly at random from the same correct samples.
Figure~\ref{fig:memorybank} shows the resulting distributions.

Across both model scales, FPS selects substantially more separated trajectories than random selection.
Even the closest FPS pair is more distant than the random selection's mean pairwise distance for 78\% of problems on Qwen2.5-0.5B-Instruct and 86\% on Qwen2.5-1.5B-Instruct.
The median distance of the closest FPS pair is approximately $1.5$--$1.6\times$ that of random selection, while the farthest FPS pairs are separated further still.
These results suggest that randomly sampled correct trajectories are often concentrated around similar reasoning paths, whereas FPS preferentially retains more diverse prototypes.

\section{Training Dynamics}\label{app:dynamic}

\begin{figure}[!t]
    \centering
    \includegraphics[width=\linewidth]{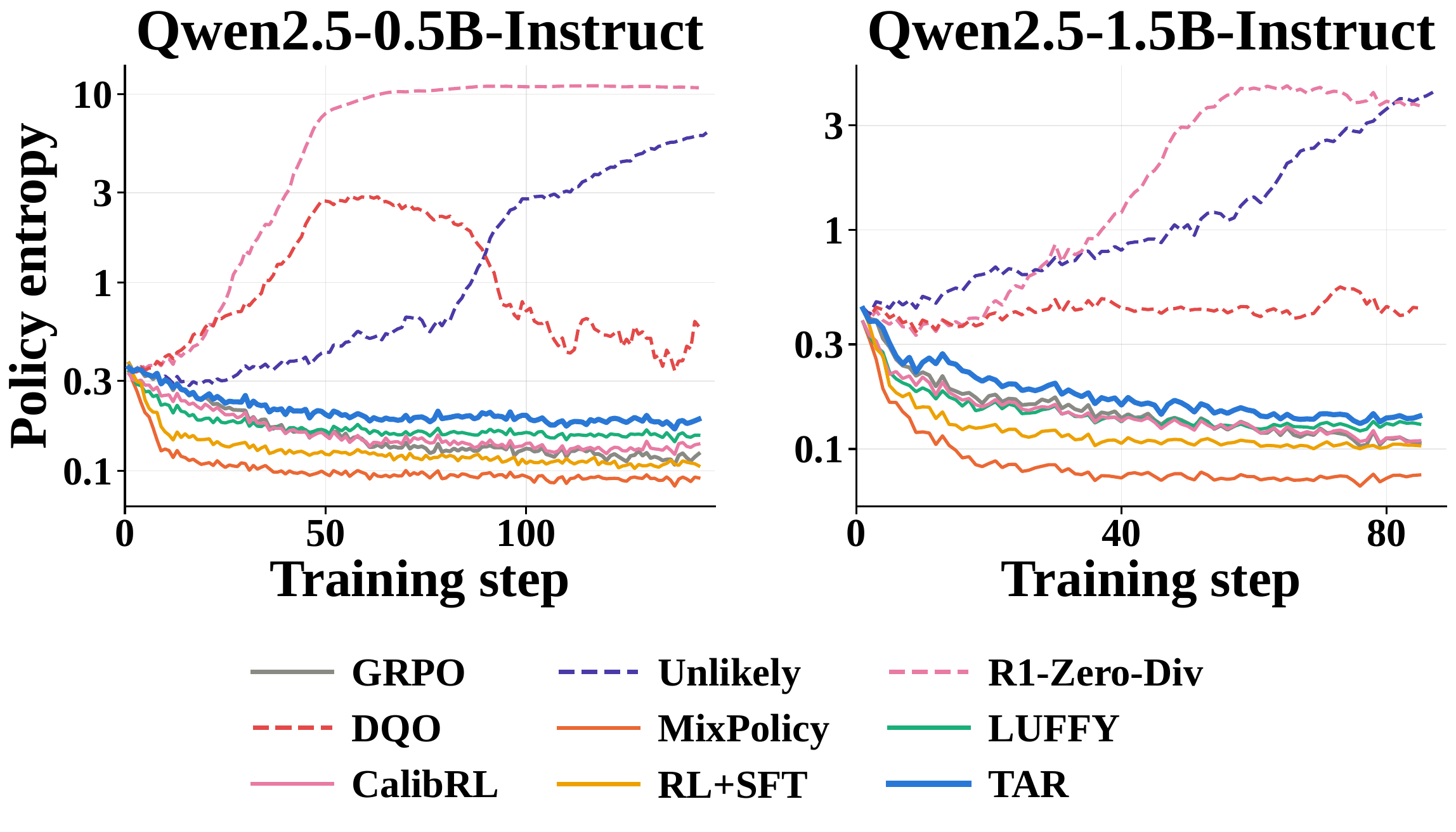}
    \caption{Training dynamics comparison in terms of generation entropy.}
    \label{fig:entropy}
\end{figure}

Figure~\ref{fig:entropy} compares policy entropy during training across different methods.
Several on-policy diversity methods, particularly R1-Zero-Div and DQO, exhibit substantial entropy growth, whereas standard GRPO and offline replay methods generally show decreasing entropy.
Despite the increased entropy, these diversity-oriented methods do not achieve corresponding improvements in reasoning performance.
In contrast, \model maintains relatively high entropy among replay-based methods while achieving the strongest average reasoning performance.
These observations further suggest that higher policy entropy does not necessarily translate into improved reasoning ability.

\section{More Experiments}\label{app:exp}

\paragraph{Diversity.} Table~\ref{tab:diversity-main} provides the full per-benchmark results on diversity corresponding to Table~\ref{tab:diversity}.

\begin{table*}[t]
\centering

\resizebox{0.9\textwidth}{!}{%
\begin{tabular}{clcccccc}
\toprule
&
\textbf{Model}
&
\textbf{GSM8K}
&
\textbf{MATH-500}
&
\textbf{AMC}
&
\textbf{College Math}
&
\textbf{Minerva}
&
\textbf{OlympiadBench}
\\
\midrule

\multirow{12}{*}{\rotatebox{90}{\textbf{Qwen-2.5-0.5B}}}
& Base
& 0.6578 / 0.4214 / 0.0424
& 0.4513 / 0.4600 / 0.0756
& 0.4891 / 0.4949 / 0.0969
& 0.4451 / 0.4679 / 0.0665
& 0.5791 / 0.5456 / 0.0989
& 0.4543 / 0.4914 / 0.1018
\\

& GRPO
& 0.4636 / 0.2823 / 0.0230
& 0.3780 / 0.3573 / 0.0523
& 0.4789 / 0.4361 / 0.0712
& 0.3806 / 0.3570 / 0.0470
& 0.5280 / 0.4683 / 0.0755
& 0.4219 / 0.4271 / 0.0755
\\

\cmidrule(lr){2-8}

& Unlikely
& 0.4117 / 0.2421 / 0.1109
& 0.3699 / 0.2900 / 0.1272
& 0.4660 / 0.3397 / 0.1441
& 0.3397 / 0.3112 / 0.1238
& 0.3940 / 0.3760 / 0.1447
& 0.3988 / 0.3426 / 0.1456
\\

& R1-Zero-Div
& 0.9913 / 0.5851 / 0.2146
& 0.9877 / 0.5202 / 0.1873
& 0.9754 / 0.4931 / 0.1874
& 0.9895 / 0.7051 / 0.1943
& 0.9838 / 0.7298 / 0.2292
& 0.9948 / 0.5294 / 0.1982
\\

& DQO
& 0.6989 / 0.5464 / 0.3579
& 0.6757 / 0.6541 / 0.4435
& 0.8300 / 0.7982 / 0.5568
& 0.5126 / 0.5041 / 0.2224
& 0.7860 / 0.7470 / 0.5532
& 0.8257 / 0.8043 / 0.6055
\\

& DPH-RL
& 0.5402 / 0.3303 / 0.0294
& 0.4081 / 0.3993 / 0.0611
& 0.4794 / 0.4552 / 0.0817
& 0.4045 / 0.3966 / 0.0544
& 0.5481 / 0.5036 / 0.0848
& 0.4445 / 0.4566 / 0.0869
\\

\cmidrule(lr){2-8}

& MixPolicy
& 0.5155 / 0.3008 / 0.0250
& 0.3827 / 0.3741 / 0.0610
& 0.4539 / 0.4354 / 0.0774
& 0.3749 / 0.3773 / 0.0528
& 0.5197 / 0.4764 / 0.0821
& 0.4087 / 0.4229 / 0.0856
\\

& LUFFY
& 0.5685 / 0.3491 / 0.0323
& 0.4038 / 0.4103 / 0.0671
& 0.4588 / 0.4690 / 0.0864
& 0.4090 / 0.4220 / 0.0587
& 0.5468 / 0.5121 / 0.0868
& 0.4301 / 0.4567 / 0.0922
\\

& CalibRL
& 0.4921 / 0.3011 / 0.0279
& 0.3854 / 0.3753 / 0.0596
& 0.4667 / 0.4521 / 0.0864
& 0.3827 / 0.3684 / 0.0506
& 0.5261 / 0.4823 / 0.0822
& 0.4229 / 0.4443 / 0.0847
\\

\cmidrule(lr){2-8}

& SFT
& 0.5761 / 0.3468 / 0.0337
& 0.4096 / 0.4018 / 0.0677
& 0.4515 / 0.4372 / 0.0860
& 0.4020 / 0.4080 / 0.0586
& 0.5395 / 0.4899 / 0.0886
& 0.4201 / 0.4351 / 0.0909
\\

& RL+SFT
& 0.5634 / 0.3422 / 0.0328
& 0.3978 / 0.3976 / 0.0663
& 0.4623 / 0.4423 / 0.0876
& 0.4008 / 0.4088 / 0.0579
& 0.5339 / 0.4991 / 0.0860
& 0.4228 / 0.4352 / 0.0911
\\

& \model
& 0.6002 / 0.3876 / 0.0383
& 0.4200 / 0.4248 / 0.0670
& 0.4687 / 0.4619 / 0.0893
& 0.4167 / 0.4327 / 0.0602
& 0.5543 / 0.5218 / 0.0901
& 0.4441 / 0.4650 / 0.0908
\\

\midrule
\midrule

\multirow{12}{*}{\rotatebox{90}{\textbf{Qwen-2.5-1.5B}}}
& Base
& 0.5886 / 0.4405 / 0.0444
& 0.4115 / 0.4648 / 0.0721
& 0.5065 / 0.5038 / 0.0950
& 0.4084 / 0.4683 / 0.0603
& 0.5527 / 0.5406 / 0.0921
& 0.4269 / 0.4830 / 0.0902
\\

& GRPO
& 0.4118 / 0.2866 / 0.0220
& 0.3403 / 0.3504 / 0.0483
& 0.4388 / 0.4392 / 0.0747
& 0.3301 / 0.3371 / 0.0398
& 0.4896 / 0.4569 / 0.0718
& 0.3976 / 0.4311 / 0.0727
\\

\cmidrule(lr){2-8}

& Unlikely
& 0.2106 / 0.3780 / 0.0753
& 0.2306 / 0.3936 / 0.0938
& 0.3321 / 0.4514 / 0.1114
& 0.2845 / 0.4284 / 0.0836
& 0.3572 / 0.4664 / 0.1131
& 0.3004 / 0.4334 / 0.1089
\\

& R1-Zero-Div
& 0.5269 / 0.4231 / 0.0365
& 0.4262 / 0.4432 / 0.0667
& 0.5217 / 0.5056 / 0.0836
& 0.4318 / 0.4206 / 0.0519
& 0.5911 / 0.5456 / 0.1210
& 0.4587 / 0.4926 / 0.0861
\\

& DQO
& 0.5481 / 0.4518 / 0.0892
& 0.5134 / 0.5609 / 0.2314
& 0.5882 / 0.5878 / 0.2980
& 0.4632 / 0.5508 / 0.1477
& 0.5950 / 0.6061 / 0.1880
& 0.5386 / 0.5822 / 0.2822
\\

& DPH-RL
& 0.4697 / 0.3246 / 0.0265
& 0.3695 / 0.3895 / 0.0568
& 0.4503 / 0.4635 / 0.0828
& 0.3574 / 0.3776 / 0.0466
& 0.5137 / 0.4854 / 0.0793
& 0.4157 / 0.4545 / 0.0821
\\

\cmidrule(lr){2-8}

& MixPolicy
& 0.4303 / 0.2967 / 0.0243
& 0.3407 / 0.3552 / 0.0533
& 0.3946 / 0.3958 / 0.0739
& 0.3224 / 0.3393 / 0.0417
& 0.4936 / 0.4628 / 0.0747
& 0.3782 / 0.4097 / 0.0741
\\

& LUFFY
& 0.4859 / 0.3447 / 0.0299
& 0.3645 / 0.3954 / 0.0582
& 0.4467 / 0.4542 / 0.0824
& 0.3516 / 0.3793 / 0.0470
& 0.5223 / 0.4911 / 0.0839
& 0.3968 / 0.4429 / 0.0803
\\

& CalibRL
& 0.4235 / 0.2912 / 0.0226
& 0.3392 / 0.3521 / 0.0490
& 0.4272 / 0.4247 / 0.0724
& 0.3308 / 0.3368 / 0.0403
& 0.4903 / 0.4541 / 0.0695
& 0.3934 / 0.4237 / 0.0713
\\

\cmidrule(lr){2-8}

& SFT
& 0.5170 / 0.3727 / 0.0394
& 0.3755 / 0.4168 / 0.0655
& 0.4354 / 0.4355 / 0.0869
& 0.3707 / 0.4152 / 0.0547
& 0.5254 / 0.4959 / 0.0864
& 0.3952 / 0.4342 / 0.0835
\\

& RL+SFT
& 0.5072 / 0.3703 / 0.0389
& 0.3777 / 0.4122 / 0.0643
& 0.4512 / 0.4381 / 0.0853
& 0.3700 / 0.4152 / 0.0545
& 0.5288 / 0.4930 / 0.0871
& 0.3989 / 0.4342 / 0.0830
\\

& \model
& 0.4986 / 0.3621 / 0.0317
& 0.3726 / 0.3988 / 0.0576
& 0.4399 / 0.4518 / 0.0831
& 0.3621 / 0.3957 / 0.0483
& 0.5293 / 0.5018 / 0.0834
& 0.4065 / 0.4478 / 0.0793
\\

\bottomrule
\end{tabular}%
}
\caption{
Diversity results on Qwen-2.5-Instruct using pass@8 rollouts.
Each cell reports Div-Eq / EAD-Distinct-N / Emb-Div.
}
\label{tab:diversity-main}

\end{table*}

\paragraph{Ablation study.} Tables~\ref{tab:reasoning-ablation-15b} and \ref{tab:diversity-ablation-15b} provide the full per-benchmark results on the ablation study corresponding to Table~\ref{tab:ablation}.

\begin{table*}[t]
\centering

\resizebox{\linewidth}{!}{%
\begin{tabular}{l|cccccc}
\toprule
\textbf{Model}
&
\textbf{GSM8K}
&
\textbf{MATH-500}
&
\textbf{AMC}
&
\textbf{College Math}
&
\textbf{Minerva}
&
\textbf{OlympiadBench}
\\
\midrule

\model
& \textbf{76.1} / 91.7
& \textbf{59.0} / \textbf{79.0}
& 30.0 / \textbf{67.5}
& 39.9 / 50.7
& \textbf{19.5} / \textbf{36.0}
& \textbf{22.4} / \textbf{41.6}
\\

\quad w/o token gate
& 74.0 / 91.8
& 58.4 / 77.4
& 30.0 / 60.0
& \textbf{40.0} / 50.4
& \textbf{19.5} / 34.2
& 20.7 / 40.4
\\

\quad w/o rollout gate
& 72.7 / \textbf{92.7}
& 53.0 / 77.0
& 30.0 / 57.5
& 37.3 / \textbf{51.7}
& 18.0 / 33.1
& 19.7 / 40.3
\\

\quad hard rollout gate
& 72.5 / \textbf{92.7}
& 53.6 / 77.0
& \textbf{35.0} / \textbf{67.5}
& 38.6 / 51.1
& 17.6 / 33.5
& 18.7 / 39.1
\\

\bottomrule
\end{tabular}%
}
\caption{Ablation study on Qwen-2.5-1.5B-Instruct. Each cell reports pass@1 / pass@8.}
\label{tab:reasoning-ablation-15b}

\end{table*}

\begin{table*}[t]
\centering

\resizebox{\linewidth}{!}{%
\begin{tabular}{l|cccccc}
\toprule
\textbf{Model}
&
\textbf{GSM8K}
&
\textbf{MATH-500}
&
\textbf{AMC}
&
\textbf{College Math}
&
\textbf{Minerva}
&
\textbf{OlympiadBench}
\\
\midrule

\model
& 0.4986 / 0.3621 / 0.0317
& 0.3726 / 0.3988 / 0.0576
& 0.4399 / 0.4518 / 0.0831
& 0.3621 / 0.3957 / 0.0483
& 0.5293 / 0.5018 / 0.0834
& 0.4065 / 0.4478 / 0.0793
\\

\quad w/o token gate
& 0.4554 / 0.3162 / 0.0250
& 0.3504 / 0.3691 / 0.0525
& 0.4321 / 0.4467 / 0.0752
& 0.3386 / 0.3530 / 0.0430
& 0.5260 / 0.4827 / 0.0825
& 0.4046 / 0.4349 / 0.0755
\\

\quad w/o rollout gate
& 0.5805 / 0.4337 / 0.0436
& 0.4139 / 0.4609 / 0.0709
& 0.4819 / 0.4901 / 0.0900
& 0.4021 / 0.4576 / 0.0591
& 0.5503 / 0.5317 / 0.0928
& 0.4293 / 0.4787 / 0.0904
\\

\quad hard rollout gate
& 0.5729 / 0.4250 / 0.0423
& 0.4113 / 0.4527 / 0.0699
& 0.4731 / 0.4862 / 0.0876
& 0.3970 / 0.4483 / 0.0576
& 0.5488 / 0.5332 / 0.0924
& 0.4251 / 0.4765 / 0.0889
\\

\bottomrule
\end{tabular}%
}
\caption{
Ablation study of diversity on Qwen-2.5-1.5B-Instruct using pass@8 rollouts.
Each cell reports Div-Eq / EAD / Emb-Div.
}
\label{tab:diversity-ablation-15b}

\end{table*}

\begin{table*}[t]
\centering

\resizebox{\textwidth}{!}{%
\begin{tabular}{clcccccc}
\toprule
& \textbf{Model}
& \textbf{GSM8K}
& \textbf{MATH-500}
& \textbf{AMC}
& \textbf{College Math}
& \textbf{Minerva}
& \textbf{OlympiadBench}
\\
\midrule

\multirow{12}{*}{\rotatebox{90}{\textbf{Qwen-2.5-0.5B}}}

& Base
& 0.5915 / 0.4463 / 0.0320
& 0.4208 / 0.4847 / 0.0488
& 0.5773 / 0.5681 / 0.0571
& 0.3877 / 0.4784 / 0.0444
& 0.5569 / 0.6017 / 0.0804
& 0.4758 / 0.5470 / 0.0745
\\

& GRPO
& 0.3818 / 0.2913 / 0.0159
& 0.3235 / 0.3567 / 0.0328
& 0.5216 / 0.4712 / 0.0419
& 0.3234 / 0.3516 / 0.0283
& 0.4934 / 0.5008 / 0.0545
& 0.4396 / 0.4724 / 0.0450
\\

\cmidrule(lr){2-8}

& Unlikely
& 0.4116 / 0.2439 / 0.1003
& 0.3490 / 0.2644 / 0.1064
& 0.5451 / 0.3641 / 0.1333
& 0.3016 / 0.2753 / 0.1093
& 0.3548 / 0.3218 / 0.1362
& 0.4226 / 0.3322 / 0.1291
\\

& R1-Zero-Div
& 0.9862 / 0.3991 / 0.1253
& 0.9934 / 0.3268 / 0.0891
& 1.0000 / 0.2523 / 0.0744
& 0.9968 / 0.4935 / 0.0840
& 0.8562 / 0.4910 / 0.0936
& 0.9891 / 0.2892 / 0.0853
\\

& DQO
& 0.5213 / 0.4181 / 0.0332
& 0.3855 / 0.4319 / 0.0487
& 0.6512 / 0.5253 / 0.0431
& 0.3714 / 0.4211 / 0.0397
& 0.4534 / 0.4830 / 0.0528
& 0.5509 / 0.6006 / 0.1010
\\

& DPH-RL
& 0.4594 / 0.3456 / 0.0217
& 0.3585 / 0.4070 / 0.0368
& 0.5611 / 0.5498 / 0.0511
& 0.3397 / 0.3931 / 0.0341
& 0.5296 / 0.5381 / 0.0567
& 0.4606 / 0.5101 / 0.0608
\\

\cmidrule(lr){2-8}

& MixPolicy
& 0.4613 / 0.3220 / 0.0175
& 0.3438 / 0.3851 / 0.0341
& 0.4924 / 0.4398 / 0.0458
& 0.3181 / 0.3755 / 0.0324
& 0.4542 / 0.4802 / 0.0647
& 0.4362 / 0.4736 / 0.0539
\\

& LUFFY
& 0.5038 / 0.3716 / 0.0242
& 0.3659 / 0.4170 / 0.0421
& 0.5497 / 0.5024 / 0.0548
& 0.3536 / 0.4262 / 0.0389
& 0.4837 / 0.5450 / 0.0679
& 0.4669 / 0.5167 / 0.0652
\\

& CalibRL
& 0.4234 / 0.3182 / 0.0201
& 0.3404 / 0.3690 / 0.0378
& 0.5630 / 0.5276 / 0.0739
& 0.3245 / 0.3600 / 0.0302
& 0.4531 / 0.4846 / 0.0557
& 0.4403 / 0.4904 / 0.0560
\\

\cmidrule(lr){2-8}

& SFT
& 0.5229 / 0.3761 / 0.0251
& 0.3730 / 0.4177 / 0.0407
& 0.6271 / 0.5689 / 0.0486
& 0.3415 / 0.4150 / 0.0382
& 0.4799 / 0.4921 / 0.0652
& 0.4428 / 0.4862 / 0.0585
\\

& RL+SFT
& 0.5067 / 0.3689 / 0.0245
& 0.3760 / 0.4200 / 0.0419
& 0.5762 / 0.5758 / 0.0556
& 0.3448 / 0.4139 / 0.0374
& 0.4794 / 0.5202 / 0.0670
& 0.4740 / 0.5373 / 0.0672
\\

& \model
& 0.5383 / 0.4058 / 0.0294
& 0.3846 / 0.4373 / 0.0435
& 0.6187 / 0.5420 / 0.0417
& 0.3556 / 0.4369 / 0.0400
& 0.5248 / 0.5685 / 0.0674
& 0.4604 / 0.5180 / 0.0609
\\

\midrule
\midrule

\multirow{12}{*}{\rotatebox{90}{\textbf{Qwen-2.5-1.5B}}}

& Base
& 0.5658 / 0.4592 / 0.0404
& 0.3900 / 0.4835 / 0.0559
& 0.5298 / 0.5675 / 0.0596
& 0.3800 / 0.4903 / 0.0469
& 0.5308 / 0.5718 / 0.0763
& 0.4483 / 0.5278 / 0.0643
\\

& GRPO
& 0.3917 / 0.3028 / 0.0193
& 0.3158 / 0.3480 / 0.0325
& 0.4346 / 0.4727 / 0.0496
& 0.2931 / 0.3364 / 0.0269
& 0.4594 / 0.4866 / 0.0511
& 0.3949 / 0.4469 / 0.0488
\\

\cmidrule(lr){2-8}

& Unlikely
& 0.2212 / 0.4017 / 0.0717
& 0.2246 / 0.4117 / 0.0797
& 0.3961 / 0.5249 / 0.0851
& 0.2554 / 0.4402 / 0.0757
& 0.3490 / 0.4485 / 0.1065
& 0.3219 / 0.4603 / 0.0888
\\

& R1-Zero-Div
& 0.5033 / 0.4424 / 0.0324
& 0.4022 / 0.4506 / 0.0471
& 0.5352 / 0.5243 / 0.0595
& 0.3929 / 0.4260 / 0.0370
& 0.5419 / 0.5748 / 0.0981
& 0.4698 / 0.5184 / 0.0565
\\

& DQO
& 0.5222 / 0.4646 / 0.0791
& 0.4406 / 0.5464 / 0.1276
& 0.5843 / 0.6150 / 0.1602
& 0.4307 / 0.5722 / 0.1215
& 0.5817 / 0.6433 / 0.1508
& 0.5142 / 0.5749 / 0.1496
\\

& DPH-RL
& 0.4417 / 0.3389 / 0.0230
& 0.3421 / 0.3908 / 0.0404
& 0.4729 / 0.5109 / 0.0561
& 0.3266 / 0.3892 / 0.0333
& 0.4787 / 0.5095 / 0.0637
& 0.4275 / 0.4818 / 0.0537
\\

\cmidrule(lr){2-8}

& MixPolicy
& 0.4067 / 0.3106 / 0.0207
& 0.3052 / 0.3466 / 0.0364
& 0.4056 / 0.4377 / 0.0496
& 0.2898 / 0.3419 / 0.0284
& 0.4379 / 0.4782 / 0.0603
& 0.3868 / 0.4442 / 0.0517
\\

& LUFFY
& 0.4627 / 0.3587 / 0.0263
& 0.3407 / 0.4006 / 0.0420
& 0.4869 / 0.4990 / 0.0577
& 0.3180 / 0.3839 / 0.0344
& 0.4659 / 0.4998 / 0.0614
& 0.4196 / 0.4761 / 0.0553
\\

& CalibRL
& 0.3996 / 0.3040 / 0.0194
& 0.3123 / 0.3450 / 0.0339
& 0.4293 / 0.4654 / 0.0432
& 0.2893 / 0.3357 / 0.0277
& 0.4426 / 0.4760 / 0.0503
& 0.4036 / 0.4425 / 0.0466
\\

\cmidrule(lr){2-8}

& SFT
& 0.4998 / 0.3912 / 0.0353
& 0.3542 / 0.4272 / 0.0498
& 0.5425 / 0.5442 / 0.0634
& 0.3505 / 0.4398 / 0.0436
& 0.5066 / 0.5259 / 0.0613
& 0.4289 / 0.4896 / 0.0597
\\

& RL+SFT
& 0.4865 / 0.3845 / 0.0349
& 0.3589 / 0.4253 / 0.0484
& 0.5232 / 0.5318 / 0.0649
& 0.3445 / 0.4379 / 0.0432
& 0.4973 / 0.5098 / 0.0603
& 0.4203 / 0.4792 / 0.0581
\\

& \model
& 0.4714 / 0.3753 / 0.0285
& 0.3491 / 0.4061 / 0.0432
& 0.5136 / 0.4996 / 0.0577
& 0.3284 / 0.4071 / 0.0363
& 0.5214 / 0.5378 / 0.0572
& 0.4328 / 0.4792 / 0.0546
\\

\bottomrule
\end{tabular}%
}
\caption{
Diversity of correct rollouts on Qwen-2.5-Instruct using pass@8 sampling.
Only problems with at least two correct responses are included.
Each cell reports Div-Eq / EAD / Emb-Div.
}
\label{tab:correct-diversity}

\end{table*}

\paragraph{Diversity on correct rollouts.} We further evaluate diversity among correct responses to examine whether the improvements observed in
Table~\ref{tab:diversity} extend to correct solutions. For each problem, we retain only correct responses
from the eight evaluation rollouts and compute diversity metrics when at least two correct responses
are available.
As shown in Table~\ref{tab:correct-diversity}, \model generally achieves higher correct-response diversity than GRPO across both model scales, following a similar trend to the unconditional diversity results. Since this filtering is performed separately for each model, the evaluated problem subsets may differ across methods.

\end{document}